%% file: main.tex
\documentclass[conference]{IEEEtran}
\IEEEoverridecommandlockouts
\usepackage{cite}
\usepackage{amsmath,amssymb,amsfonts}
\usepackage{algorithmic}
\usepackage{graphicx}
\usepackage{textcomp}
\usepackage{xcolor}
\def\BibTeX{{\rm B\kern-.05em{\sc i\kern-.025em b}\kern-.08em
    T\kern-.1667em\lower.7ex\hbox{E}\kern-.125emX}}

\newtheorem{theorem}{Theorem}[section]
\newtheorem{problem}[theorem]{Problem}
\newtheorem{example}[theorem]{Example}
\newtheorem{definition}[theorem]{Definition}
\newtheorem{proof}[theorem]{Proof}
\newtheorem{proposition}[theorem]{Proposition}

\newcommand{\minK}{\ensuremath{\tilde{k}}}

\usepackage{tikz}
\usetikzlibrary{positioning}
\usetikzlibrary{fit,calc, backgrounds}
\usepackage{paralist}
\usepackage{enumitem}
\usepackage{color, colortbl}
\usepackage{multirow}
\usepackage{makecell}
\usepackage{url}

\usepackage{xcolor}
\usepackage{subcaption}
\usepackage[vlined,ruled,linesnumbered]{algorithm2e}
\let\oldnl\nl
\newcommand{\nonl}{\renewcommand{\nl}{\let\nl\oldnl}}

\SetCommentSty{mycommfont}

\newcommand{\yuval}[1]{}
\newcommand{\jinyang}[1]{}
\newcommand{\jag}[1]{}

\definecolor{ForestGreen}{RGB}{34,139,34}
\newcommand{\rev}[1]{{#1}}

\newcommand{\fullVerOnly}[1]{{#1}}
\newcommand{\notFullVerOnly}[1]{}

\begin{document}

\title{Detection of Groups with \\Biased Representation in Ranking
}

\author{\IEEEauthorblockN{Jinyang Li}
\IEEEauthorblockA{
\textit{University of Michigan}\\
jinyli@umich.edu}
\and
\IEEEauthorblockN{Yuval Moskovitch}
\IEEEauthorblockA{
\textit{Ben Gurion University of the Negev}\\
yuvalmos@bgu.ac.il}
\and
\IEEEauthorblockN{H. V. Jagadish}
\IEEEauthorblockA{
\textit{University of Michigan}\\
jag@umich.edu}
}

\maketitle

\begin{abstract}
Real-life tools for decision-making in many critical domains are based on ranking results. With the increasing awareness of algorithmic fairness, recent works have presented measures for fairness in ranking. Many of those definitions consider the representation of different ``protected groups'', in the top-$k$ ranked items, for any reasonable $k$. 
Given the protected groups, confirming algorithmic fairness is a simple task. However, the groups' definitions may be unknown in advance.

In this paper, we study the problem of detecting groups with biased representation in the top-$k$ ranked items, eliminating the need to pre-define 
protected groups.
The number of such groups possible can be exponential, making the problem hard. 
We propose efficient search algorithms  for two different fairness measures: global representation bounds, and proportional representation.
Then we propose a method to explain the bias in the representations of groups utilizing the notion of Shapley values. We conclude with an experimental study, showing the scalability of our approach and demonstrating the usefulness of the proposed algorithms.
\end{abstract}


\input{intro}
\input{prelim}

\input{problem_def}
\input{ranking}

\input{explanations}

\input{exp}

\input{related}

\input{conc}

\section*{Acknowledgement}
We would like to thank the anonymous reviewers for their comments and suggestions. This research is supported in part by NSF under grants 1741022, 1934565 and 2106176.
\clearpage

\bibliographystyle{plain}
\bibliography{bibtex}

\end{document}

%% file: intro.tex
\section{Introduction}

Ranking is a commonly used operation in a wide range of application domains, for example, in presenting results on a web search engine~\cite{BrinP98}, establishing credit scores~\cite{berg2020rise}, school admission~\cite{peskun2007effectiveness} and hiring~\cite{GeyikAK19}. While convenient and useful, these tools can be biased. As a result, they may affect decision-making in undesirable ways and can even impact human life~\cite{ChenMHW18, barocas2016big, AltenburgerDFAH17}. This problem has drawn much attention from the research community, and a line of recent works has focused on measuring and mitigating bias and unfairness in ranking~\cite{AsudehJS019, YangS17, SinghJ18, CelisSV18, GeyikAK19, KuhlmanVR19, ZehlikeB0HMB17}.




The notion of algorithmic fairness was studied extensively for a broad class of models~\cite{rankingFairness, mitchell2021algorithmic}. Fairness measures typically refer to a given ``protected group'' in the data, which is defined based on the values of some sensitive attributes (e.g., gender, race, age, or combinations thereof), usually based on the societal history of discrimination. Analyzing the fairness measure of a system with respect to the given group is a simple task. However, ``non-standard'' protected groups cannot always be specified in advance, and such groups may be overlooked when examining the performance of a system.

For example, a model developed to assign grades to students (in place of exams that were canceled due to the COVID-19 pandemic) was shown to be biased against high-achieving students from poor school districts\footnote{\url{https://www.nytimes.com/2020/09/08/opinion/international-baccalaureate-algorithm-grades.html}}. For instance, students from low-income families were predicted to fail the Spanish exam, even when they were native Spanish speakers. In this case, the model was discriminating against Hispanic students from poor school districts. 
A primary source of bias was the use of historical exam results of each school to predict student performance. However, using the school (identified by school ID) to define the protected group is not an intuitive choice, and so may not have been considered. Moreover, even if we consider the group of Hispanic students as a protected group, we may not find any fairness issues, since this subgroup of students is only a small fraction of all Hispanic students.

In this paper we study the problem of detecting groups that are treated unfairly by a ranking algorithm. In other words, we want to let the data speak to (potential) unfairness, without requiring a human modeler to identify protected attributes ahead of time. Following fairness definitions presented in the literature on fairness in ranking (see e.g.,~\cite{CelisSV18, YangS17, rankingFairness}), we consider group representation in the top-$k$ ranked items for any $k$ in a reasonable range as a measure of fairness.

Recent works have studied the problem of automatically detecting ``problematic'' or biased subgroups in the data without the need to specify the protected attributes a priori~\cite{ChungKPTW20, cabrera2019fairvis, pastor2021looking, jin2020mithracoverage, LiMJ21}. However, these works considered only classification models. In~\cite{pastor2021identifying} the authors of~\cite{pastor2021looking} extend their framework to consider ranking as well. In contrast to our work, which builds on fairness measures for ranking from the literature, they use the notion of divergence to measure performance differences among data subgroups. This difference leads to differences in the result sets returned by each method (see Section~\ref{sec:comp} for more details). 

We next outline our main contributions.

\paragraph*{Problem formulation} We formally define the problem of detecting groups with biased representation in the top-$k$ ranked items for a given ranking algorithm $R$, a dataset $D$, and a range of possible $k$'s. Groups are defined using value assignments to a set of attributes we denote as \emph{patterns} (see Section~\ref{sec:prelim}).  To provide concise and meaningful results we use a threshold on the returned groups' size and report only groups that are not subsumed by any other group in the result set (referred to as the \emph{most general patterns}, see Section~\ref{sec:problem}). 
We start with the fundamental definition of~\cite{CelisSV18}, which uses upper and lower bounds to restrict the number of tuples in the top-$k$ from different groups in the data.  The goal is to report groups such that their representation (i.e., number of tuples) in the top-$k$ does not lie within the given bounds for a given range of possible $k$'s. We refer to this problem as the \emph{global bounds representation bias} problem. We then consider the prominent class of fairness measures utilizing proportional representation (see, e.g.,~\cite{YangS17}). Intuitively, the representation of each group in the top-$k$ should be proportional to its size in the data. Using this notion we define the \emph{proportional representation bias} problem. We show that no polynomial algorithm exists to solve either problem.

\paragraph*{Detection of groups with biased representation} We present algorithms for the problem of finding the set of all substantial groups (in terms of their size in the data, and their subsumption in other groups) with biased representation in the top-$k$ ranked items. We first present a simple baseline solution that utilizes the notion of pattern graph presented in \cite{AsudehJJ19}. We show how to traverse the graph in a top-down fashion in order to find groups with biased representation. This search is then applied repeatedly for each $k$ in the given range. Bearing in mind the complexity of the problem, we focus on optimizing the search. Our optimized solutions rely on the fact that the set of top-$k$ and top-$(k+1)$ tuples differ by a single tuple. As a result, the search spaces for succeeding $k$ values are typically very similar. The optimized solutions utilize this observation to avoid parts of the search tree.

\paragraph*{Result analysis} Given a group with biased representation in the top-$k$ ranked items, an analyst may wish to understand the cause of bias. To this end, we propose a method that harnesses the notion of Shapley values to identify attributes that significantly affect the ranking of the detected group. To analyze the difference between the detected group and top-$k$ ranked tuples, we visualize the value distribution of such attributes. Shapley values have been used to provide similar explanations  for regression and classification models. Our novelty is in developing a corresponding method for the ranking problem. 

\paragraph*{Experimental study} We complement our algorithmic development with an extensive experimental study. We evaluate the performance and properties of the algorithms, i.e., the scalability and parameter setting effect. We examine the effect of the number of attributes, groups' size threshold, and range of $k$, using three real-world datasets. Our results show the applicability of our solution in practice, despite the theoretical complexity of the problems, and the usefulness of the optimized algorithms compared to the baseline solution. We then experimentally demonstrate the usefulness of our approach for results analysis. Finally, we compare the result of our algorithms to the results of the method proposed in~\cite{pastor2021identifying} through a case study.

\paragraph*{Paper organization} The rest of the paper is organized as follows. We present the necessary preliminaries for our problem definition in Section~\ref{sec:prelim}. Then in Section~\ref{sec:problem} we formally define the problems of detecting groups with biased representation in the data and prove their hardness. Our solutions is presented in Section~\ref{sec:algo}. In Section~\ref{sec:explanations} we introduce a method for analyzing and explaining the results of our algorithms. We describe our experimental evaluation in Section~\ref{sec:exp}, overview related work in Section~\ref{sec:related}, and conclude in Section~\ref{sec:conc}.

%% file: prelim.tex
\section{preliminaries}
\label{sec:prelim}


We next provide the necessary background on the notion of patterns to represent data groups and the concept of fairness in ranking.
We will use the following example as our running example to demonstrate the ideas presented in the paper.

\begin{example}\label{ex:running}
The Student Performance Data Set~\cite{cortez2008using} contains information
from two Portuguese secondary schools in the Alentejo region of Portugal, Gabriel Pereira (GP) and Mousinho da Silveira (MS). The data was collected during the 2005-2006 school year and it contains the performance of 1044 students in the Math and the Portuguese language exams, along with  demographic, social, and school-related information. Figure~\ref{fig:dataset} depicts a sample from the data with the attributes: gender, school, address (urban or rural), and failures (number of past class failures). The grade attribute depicts the students' grades on a scale of $0-20$. Consider an excellence student program committee that wishes to select students for a scholarship based on their academic achievements. To this end, they use a ranking algorithm $R$ to rank students by their grades. In the case of similar grades, students with fewer failures are ranked higher. The scholars' list is publicly announced, and should be diverse and inclusive.
\end{example}

\subsection{Data Groups}



We assume the data is represented using a single relational database, and that the relation's attribute values used for group definitions are categorical. To include attribute values drawn from a continuous domain in the group definition, we render them categorical by bucketizing them into ranges: very commonly done in practice to present aggregate results. We use the notion of patterns, value assignments to a set of attributes, to define groups in the data~\cite{AsudehJJ19}.


	\begin{definition}[Patterns]
		Let $D$ be a database with categorical attributes $\mathcal{A}=\{A_1,\ldots, A_n\}$ and let $Dom(A_i)$ be the active domain of $A_i$ for $i \in [1..n]$. A \emph{pattern} $p$ over $D$ is the set of $\{A_{i_1} = a_1, \ldots, A_{i_k} = a_k\}$ where $\{A_{i_1},\ldots, A_{i_k}\}\subseteq \mathcal{A}$ and $a_j\in Dom(A_{i_j})$ for each $A_{i_j}$ in $p$. We use $Attr(p)$ to denote the set of attributes in $p$.

	\end{definition}
	
	We say that a tuple $t\in D$ \emph{satisfies} a pattern $p$ if $t.A_i = a_i$ for each $A_i \in Attr(p)$. The \emph{size} $s_D(p)$ of a pattern $p$ is then the number of tuples in $D$ that satisfy~$p$.  Given a ranking algorithm $R$ we use $R^k(D)$ to denote the top-$k$ ranked items in $D$. Finally, we use  $s_{R^k(D)}(p)$ to denote the size of $p$ in~$R^k(D)$.
	
		\begin{example}
Consider the dataset given in Figure~\ref{fig:dataset}. $p = $ \{School = GP\}, is an example of a pattern. Tuples 3, 4, 7, 8, 12, 13, 15, and 16 satisfy $p$ and thus $s_D(p)= 8$. For the ranking algorithm $R$ whose results are depicted in the Rank column, we have $s_{R^5(D)}(p) = 1$, since only one tuple in the top-$5$ ranked items satisfies $p$.
\end{example}
	


\begin{figure}
    \centering
    \footnotesize
    \begin{tabular}{|c|c|c|c|c|c||c|}
    \hline
         \# & Gender & School & Address & Failures & Grade  & Rank\\
         \hline
         1& F & MS & R & 1 &  11    & 8\\
         \rowcolor{blue!15}2& M & MS & R & 1 &  15  & 3\\
          3& M & GP & U & 1 &  8  & 10\\ 
          4& M & GP & U & 2 &  4  & 16\\
          \rowcolor{blue!15} 5& M & MS & R & 0 &  19  &  2 \\
          6& F & MS & U & 1 &  4  &  15\\
          7& F & GP & R & 1 &  7  &  11\\
          8& M & GP & R & 1 &  6   & 13 \\
          \rowcolor{blue!15}9& F & MS & R & 0 &  14   & 4\\
          10& F & MS & R & 2 &  7    & 12\\
          11& M & MS & R & 2 &  13   & 6 \\
          \rowcolor{blue!15}12& F & GP & U & 0 &  20   & 1 \\
          13& F & GP & U & 2 &  12  & 7\\
          \rowcolor{blue!15}14& M & MS & U & 1 &  13   & 5 \\
          15& F & GP & U & 1 &  5   &  14\\
          16& M & GP & U & 0 &  9  & 9\\
         \hline
    \end{tabular}
    \caption{Students’ data. The Rank column depicts their ranking based on the grade and number of past failures. The top-$5$ ranked students are highlighted}
    \vspace{-2mm}
    \label{fig:dataset}
\end{figure}

\subsection{Fairness measure}

The problem of fairness in ranking was studied in a line of works (see~\cite{rankingFairness} for a survey). A fundamental definition, presented in~\cite{CelisSV18}, uses constraints over the representations of different groups in the top-$k$ ranked items. They use an upper bound $U_{kl}$ and a lower bound $L_{kl}$ over the number of items with the property $l$ (i.e., a protected group) in the top-$k$ positions of the ranking. Then a ranking algorithm is fair by the definition of~\cite{CelisSV18}, if the number of selected items from the protected group in the top-$k$ lies within the given boundaries.

\begin{example}\label{ex:fairness_ranking}
Consider again the dataset given in Figure~\ref{fig:dataset} and the ranker whose result is presented in the Rank column.
Consider a lower bound of 2 over the number of students from each school for $k = 5$ (i.e., $L_{5,school=MS} = L_{5,school=GP} = 2$). In this case, among the top-$5$ students, only one is from the GP school, thus the ranker does not satisfy the constraints.
\end{example}

Another prominent class of fairness measure in ranking considers the proportional representation of different groups in the top-$k$ ranked items (see, e.g~\cite{YangS17}). Intuitively, these definitions can be seen as variants of the definition of~\cite{CelisSV18} such that for each group $g$, and each $k$, the bounds on the number of occurrences of items from $g$ in the top-$k$ ranked items are defined with respect to the size of $g$ in the dataset.

\begin{example}
Continuing Example~\ref{ex:fairness_ranking}, the total number of students from each school (MS and GP) is 8. The total dataset size is 16, thus a proportionate representation of each school in the top-$5$ items should be roughly $5\cdot\frac{8}{16} \approx 2$. 
\end{example}

%% file: problem_def.tex
\section{Problem definition}\label{sec:problem}


 Our goal is to detect groups with biased representation in the top-$k$ ranked items for a given ranking algorithm $R$, dataset $D$, and a range of $k$'s. 
 \rev{We define our problem by harnessing fairness measures for ranking algorithms from the literature. In particular, we present two problem definitions utilizing prominent fairness measures, both considering the representation of different groups in the top-$k$ ranked items for different values of $k$. 
 Intuitively, accounting for a range of $k$'s ensures that the ranking is fair for any \emph{position} in the ranking. For instance, if the top-$10$ items consist of $5$ students with an urban address and $5$ students with a rural address, but the students with urban addresses are in the $1$-$5$ positions of the ranking, the output may seem ``fair'' if we are only interested in selecting the top-$10$ students, but if the positions within the top-$10$ are also important (e.g., position in the ranking affects an award amount), then clearly this ranking is unfair.}



 The first definition simply utilizes the definition of~\cite{CelisSV18}.
 The fairness definition of~\cite{CelisSV18} restricts the count of different groups (i.e., patterns) in the top-$k$ using upper and lower bounds. According to this definition, the result is biased either when the size of a pattern $l$ exceeds the upper bound $U_{kl}$ or falls below the lower bound $L_{kl}$ among the top-$k$ tuples for some $k$. We eliminate the requirement to define $l$ in advance and use $U_{k}$ and $L_{k}$ to denote the upper and lower bound respectively, on every pattern in the top-$k$ ranked tuples of a given ranking algorithm.

  We say that a group has a biased representation in the output of a ranking algorithm $R$, if its size in the top-$k$ ranked items by $R$ does not lie within the given bounds for any $k$ in a given range of possible $k$'s. Intuitively, we wish to avoid reporting  ``very specific" descriptions of groups and provide the user with a concise set of properties that characterize meaningful groups (in terms of their size) that have biased representation. To this end, we present the notion of \emph{most general} patterns.
  Given the bounds over the group's representation in the top-$k$ ranked items, we say that a pattern $p$ is the most general pattern with biased representation, if $p$ is used to represent a group with inadequate representation by the given bounds, and $\forall p'\subsetneq p$, the count of $p'$ in $R^k(D)$ lies within the given boundaries. We are now ready to formally define our problem.
  
  \begin{problem}[Global Bounds Representation  Bias]\label{problem:rankingBounds}
	Given a database $D$, a ranking algorithm $R$, a size threshold $\tau_s$\footnote{We use an absolute value as the size threshold. Equivalently it may be defined as a fraction of the dataset size.}, a range $[k_{min}, k_{max}]$, and lower bounds $L_k$ and upper bounds $U_k$ for each $k_{min}\leqslant k \leqslant k_{max}$, find for each $k_{min}\leqslant k\leqslant k_{max}$, all most general patterns $p$ with size $\geqslant \tau_s$ such that $s_{R^k(D)}(p) < L_k$ or $s_{R^k(D)}(p) > U_k$.
\end{problem}

  Note that the ranking algorithm is treated as a black box, making the problem to be model agnostic. Following the line of work on proportional representation, we consider another problem definition by refining the above definition to account for the groups' sizes in the dataset. Intuitively, the number of items from each group in the top-$k$ ranked items should be proportionate to the group's representation in the data.
  

    \begin{problem}[Proportional Representation Bias]\label{problem:ranking:proportional}
	Given a database $D$, a ranking algorithm $R$, a size threshold $\tau_s$ and a range $[k_{min}, k_{max}]$, find for each $k_{min}\leqslant k\leqslant k_{max}$, all most general patterns $p$ with size $\geqslant \tau_s$ such that $s_{R^k(D)}(p) < \alpha\cdot s_D(p)\frac{k}{|D|}$  or $s_{R^k(D)}(p) > \beta\cdot s_D(p)\frac{k}{|D|}$ for $\alpha < \beta \in \mathbb{R}$.
\end{problem}


Proportional representation gives the user an intuitive bounds definition. However, the definition of global representation allows the user to actively control the bounds over the representation of different groups in the top-$k$ ranked items, even if their representation in the overall data is low/high. For instance, consider a ranking algorithm for job applicants in fields that are dominated by men (e.g., tech companies). If the company wishes to increase the representation of women they hire but their application number is low and only the top-$k$ ranked applicants are invited for an interview, by using proportional representation, their number in the top-$k$, and as a result, the number of women invited to an interview, would be low as well. \rev{
The fundamental definition of~\cite{CelisSV18} allows the user to define bounds over the representation of the protected groups in the data for different values of $k$. Following this definition, we defined the global bounds representation problem, which assumes no prior information regarding the identity of the protected groups and aims to identify \emph{all} groups with biased representation. Note that our goal is to report \emph{only} the most general patterns (groups), providing a concise description of these groups.}

While there are typically far fewer most general patterns with biased representation than the set of all patterns with biased representation, in the worst case, their numbers can be exponential. 
 This is true even if we consider only the lower bounds (e.g., if $U_k = |D|$).

\begin{theorem}
\label{prop:nopolyRanking}
Given a dataset $D$ and a ranking algorithm $R$, no polynomial time algorithm can guarantee the enumeration of the set of all most general patterns with biased representation in the result of $R$ on $D$.
\end{theorem}

\notFullVerOnly{The proof is by construction, details are in~\cite{full}.}

\fullVerOnly{
\begin{figure}[]
	    \centering
	   \footnotesize
	    \begin{tabular}{|c|c|c|c|c|c||c|}
	         \hline
	          & $A_1$ & $A_2$  & $\cdots$ & $A_{n-1}$ & $A_n$ & Rank \\
	         \hline
	          $t_1$ & 1 & 0  & $\cdots$ & 0 & 0 &1\\
	         \hline
	         $t_2$ & 0 & 1  & $\cdots$ & 0 & 0 &2\\
	         \hline
	         $\vdots$ & $\vdots$  & $\vdots$ & $\ddots$ & $\vdots$ & $\vdots$ & $\vdots$\\
	         \hline
	         $t_n$ & 0 & 0 & $\cdots$ & 0 & 1 & $n$\\
	         \hline
	          $t_{n+1}$ & 0 & 0  & $\cdots$ & 0 & 0& $n+1$ \\
	         \hline
	    \end{tabular}
	    \caption{Dataset $D$ in the proof of Theorem \ref{prop:nopolyRanking}}
	    \label{tab:proof}
	    \vspace{0.2cm}
\end{figure}

\begin{proof}
We prove the theorem by construction. Consider a dataset $D$ with $n$  (assuming $n \geq 2$) binary attributes $\{A_1\ldots, A_n\}$ and $n+1$ tuples $t_1, \dots, t_n, t_{n+1}$ as shown in Figure~\ref{tab:proof}. 
I.e., $\forall i \in [1,n], t_i[A_i] = 1$, and $\forall j \neq i, t_i[A_j]=0$.
All the attributes of $t_{n+1}$ are zero. Let $R$ be a ranking algorithm such that the top-$k$ tuples in $D$ are the tuples $t_1,\ldots,t_k$ in Figure~\ref{tab:proof}. Let $k_{min}=k_{max}=n$, $L_k=\frac{n}{2}+1$ for Problem \ref{problem:rankingBounds} (global representation bounds), and $\alpha=\frac{n+3}{n+4}$ for Problem \ref{problem:ranking:proportional} (proportional representation), for some $n \geq 2$.

Consider a pattern $p$ with $m \leq n$ attributes $A_i$ with the value assignment $A_i=0$. Let $I$ be the set of indices of those attributes.
Among $t_1, \cdots, t_n$, the size of $p$ is $n-m$, since $\forall i \in I, t_i[i] = 1, t_i$ does not satisfy $p$, and all the other tuples satisfy $p$.
Let $m=\frac{n}{2}$, thus the size of $p$ in the top-$k$ ranked items is $s_{R^k(D)}(p)=\frac{n}{2} < L_k=\frac{n}{2}+1$ in the case of global representation bounds.
For proportional representation, we have $s_D(p) = \frac{n}{2}+1$ since $t_{n+1}$ also satisfies $p$. And we get $s_{R^k(D)}(p)=\frac{n}{2} < \alpha \cdot s_D(p) \cdot \frac{k}{|D|} = \frac{n+3}{n+4} (\frac{n}{2}+1) \frac{n}{n+1} = \frac{n}{2} \frac{(n+3)(n+2)}{(n+4)(n+1)}$.
To show $p$ is a most general pattern to report, we examine the parents of $p$, $p'$ which has $m-1$ attribute with the value assignment $0$.
For global representation bounds, we have $s_{R^k(D)}(p') = n - m + 1 = \frac{n}{2} + 1 = L_k$.
For proportional representation, we have $s_{R^k(D)}(p')=\frac{n}{2} + 1 > \alpha \cdot s_D(p') \cdot \frac{k}{|D|} = \frac{n+3}{n+4} (\frac{n}{2}+2) \frac{n}{n+1} = \frac{n+2}{2} \frac{(n+3)n}{(n+1)(n+2)}$.
As a result, every pattern with $\frac{n}{2}$ attribute assigned to 0 should be in the result set.
The number of such patterns is $\binom{n}{n/2} > \sqrt{2}^{n}$, which is exponential.
Therefore, any algorithm enumerating these patterns is exponential.
\end{proof}

}

  
 \paragraph*{Upper bounds}
  The notion of the most general patterns is motivated by the utility of the information they provide. For example, in the case of global representation bounds, if the number of females in the top-$k$ is less than the lower bound, then clearly the number of black females is below the bound. Unlike the most general patterns for the lower bound, in the case of the upper bound, the most specific patterns are more informative.
 For instance, if the number of black females is above the upper bound, then so is the number of blacks and the number of females. A plausible problem definition may account for the \emph{most specific substantial patterns}. Analogously to the definition of the most general patterns, a pattern $p$ is a \emph{most specific substantial pattern} if the size of $p$ is above a given threshold $\tau_s$ and for every pattern $p'$ such that $p \subsetneq p'$, the size of $p'$ is below the threshold $\tau_s$. The goal is then to find the most general patterns that do not satisfy the lower bound and the most specific substantial patterns that exceed the upper bound. For ease of presentation, in the rest of the paper, we will focus on the solution for Problems~\ref{problem:rankingBounds} and~\ref{problem:ranking:proportional} considering only the lower bounds. We note that 
  our solutions can be adjusted to support such problem definition (and other definitions such as most general for upper bound, and the most specific for lower bound). 

\rev{While Theorem~\ref{prop:nopolyRanking} indicates that the number of most general groups with biased representation may be exponential, our experimental evaluation shows that, in practice, their number is significantly lower. In $97.58\%$ of the times, the number of the reported groups was less than $100$. We note that presenting a large number of results may be overwhelming to the user. A user-friendly interface would organize the output by $k$ value and rank the groups by their overall size in the data or by the bias in their representation (the difference between the required representation and the actual representation). }

%% file: ranking.tex
\section{Detecting Groups with Biased Representation}
\label{sec:algo}

We next present our algorithms for detecting groups with biased representation as defined in Problems~\ref{problem:rankingBounds} and~\ref{problem:ranking:proportional}. We start with a simple solution that can be used to detect groups based on both of the problem definitions. We then present two optimized algorithms, designed to optimize the search for each of the problems.




\subsection{Iterative Top-Down Search (Baseline Solution)}\label{sec:top-down}

The first, simple solution, utilizes the algorithm presented in~\cite{AsudehJJ19} to traverse the set of possible patterns, starting with the most general ones, and compute the representation of each group in the top-$k$ ranked items in the data (for each $k$ in the given range). This is done using the notion of \emph{pattern graph}~\cite{AsudehJJ19}. Briefly, the nodes in the graph are the set of all possible patterns, and there is an edge between a pair of patterns $p$ and $p'$ if $p\subset p'$ and $p'$ can be obtained from $p$ by adding a single attribute value pair. In this case, we say that $p$ ($p'$) is a parent (child) of $p'$ ($p$). 
As shown in~\cite{AsudehJJ19}, the pattern graph can be traversed in a top-down fashion, while visiting each pattern at most once. Namely, traversing a spanning tree of the pattern graph, which we denote as the \emph{search tree}, and formally define as follows.
\begin{definition}\label{def:children}
 Let $D$ be a dataset with categorical attributes $\mathcal{A}=\{A_1,\ldots, A_n\}$. We assume attributes are ordered, and for a given set of attributes $S\subseteq \mathcal{A}$ we use $idx(S)$ to denote the  maximal index value of all attributes in $S$. Let $p$ be a node in the pattern graph of $D$. The children of $p$ in the search tree $T$ are $p'$ such that $p'$ is a child of $p$ in the pattern graph, and $idx(Attr(p')\setminus Attr(p))> idx(Attr(p))$. Namely, $p'$ is obtained from $p$ by adding a single attribute value pair such that the index of the newly added attribute is larger than the maximal index value of attributes in $p$.
\end{definition}

\begin{figure}
    \centering
    
    \begin{tikzpicture}[every node/.style={rounded corners, font=\footnotesize,draw, scale=0.9}, every path/.style={draw}]
\node (root) at (0,0) {\{\}};
\node (g=f) at (-2,-0.8) {\{G=F\}};
\node (s=gp) at (2,-0.8) {\{S=GP\}};
\node (s=ms) at (-0.8,-0.8) {\{S=MS\}};
\node (g=m) at (0.8,-0.8) {\{G=M\}};
\node[draw=none] at (0,-0.8) {$\ldots$};
\node (f_gp) at (-1,-1.6) {\{G=F, S=GP\}};
\node (f_ms) at (-3,-1.6) {\{G=F, S=MS\}};
\node (m_gp) at (3,-1.6) {\{G=M, S=GP\}};
\node (m_ms) at (1,-1.6) {\{G=M, S=MS\}};

\path[draw] (root)--(g=f);
\path[draw] (root)--(g=m);
\path[draw] (root)--(s=gp);
\path[draw] (root)--(s=ms);

\path (g=f)--(f_gp);
\path (g=f)--(f_ms);
\path (g=m)--(m_gp);
\path (g=m)--(m_ms);
\path[dashed] (s=gp)--(f_gp);
\path[dashed] (s=ms)--(f_ms);
\path[dashed] (s=gp)--(m_gp);
\path[dashed] (s=ms)--(m_ms);
\end{tikzpicture}

    \caption{Part of the pattern graph for the running example. Edges of the search tree are marked with solid lines.}
    \vspace{-2mm}
    \label{fig:graph}
\end{figure}
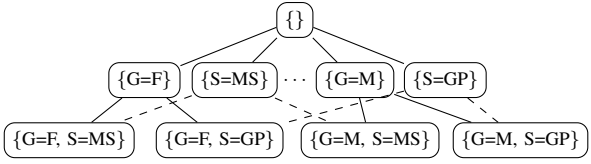

\begin{example}
A part of the pattern graph for the dataset of the running example is shown in Figure~\ref{fig:graph}, where $G$ and $S$ are used as shorthands for Gender and School respectively. The pattern \{G=F, S=GP\} is a child node of the patterns \{G=F\} and \{S=GP\} in the pattern graph, however, in the search tree it is only a child of \{G=F\}. This is because $idx(\text{\{G\}}) = 1 < 2= idx(\text{\{G, S\}})\setminus\text{\{G\}})$ whereas $idx(\text{\{S\}}) = 2 > 1= idx(\text{\{G, S\}}\setminus\text{\{S\}})$.
\end{example}

\begin{algorithm}
\footnotesize
	\DontPrintSemicolon
	\SetKwInOut{Input}{input}\SetKwInOut{Output}{output}

	\Input{A dataset $D$, a ranking algorithm $R$, a size threshold $\tau_s$, $k$ and lower bound $L_k$
	}
	\Output{$Res= \{p_1, \ldots, p_n\}$ where $\forall p_i\in Res$ $s_D(p_i) \geq \tau_s$ and $p_i$ is a most general pattern with  $s_{R^k(D)}(p) < L_k$} 
		\tcc{For proportional bounds $\alpha$ is given as input and $L_k = \alpha\cdot s_D(p)\frac{k}{|D|}$}
	\BlankLine

	\SetKwFunction{patternCount}{\texttt{patternSize}}
	\SetKwFunction{generateChildren}{\texttt{generateChildren}}
	\SetKwFunction{searchFromNode}{\texttt{searchFromNode}}
	\SetKwFunction{update}{\texttt{update}}
	\LinesNumbered
	$Res \gets \emptyset$  \label{line:initP}\\
	 $\mathcal{S} \gets \{\generateChildren(\{\})\}$  \label{line:initS}\\
	    \While{$\mathcal{S}$ is not empty}
	    { \label{line:startLoop}
	        $p = \mathcal{S}.pop()$ \label{line:pop}\\
	            \If{ \patternCount$(p, D)$ $ > \tau_s $\label{line:ifSize}}
	            {
	                $top$-$k\_c \gets$ \patternCount$(p,R^k(D))$ \label{line:countTop}\\
	                \tcc{For proportional bounds use $top$-$k$\_$c < \alpha\cdot s_D(p)\frac{k}{|D|}$}
	                \If{$top$-$k\_c < L_{k_{}}$  \label{line:ifUnfair}}
	                {
	                   \update($Res, p$) \label{line:updateRes}
	                }
	                \Else 
        	        { \label{line:else}
        	            $\mathcal{S}.push(\generateChildren(p))$ \label{line:updateS}\\
        	        }
                }
	    }\label{line:endLoop}
	    \Return $Res$\label{line:RprocReturn}
	
	\caption{Top-down search}
 \label{alg:top-down}
\end{algorithm}

Algorithm \ref{alg:top-down} detects patterns with adequate size in the data (namely above a given threshold $\tau_s$) and low representation (less than $L_k$), in the top-$k$ ranked items for a given dataset $D$, a ranking algorithm $R$ and a given $k$. It traverses the search tree of the pattern graph (top-down), using a queue $\mathcal{S}$, and maintains the set of identified patterns with $s_{R^k(D)} < L_k$. First it initializes the result set $Res$ to $\emptyset$ (line~\ref{line:initP}) and sets $\mathcal{S}$ to contain the children of the most general (empty) pattern (line~\ref{line:initS}). While the queue $\mathcal{S}$ is not empty (lines \ref{line:startLoop}~--~\ref{line:endLoop}), the algorithm extracts the first pattern in the queue $p$ (line~\ref{line:pop}), and computes its size in $D$. If it is greater than $\tau_s$ (line~\ref{line:ifSize}), the size of $p$ in $R^k(D)$, $s_{R^k(D)}$, is computed (line~\ref{line:countTop}). If $s_{R^k(D)}$ is bellow the lower bound $L_k$ (line~\ref{line:ifUnfair}), $Res$ is updated  using the procedure \texttt{update} (line~\ref{line:updateRes}), that checks whether any ancestor of $p$ in the pattern graph is already in $Res$ (this is possible since the algorithm traverses the search tree and not the patterns graph). Otherwise (line~\ref{line:else}), $s_{R^k(D)}$ exceeds the lower bound $L_k$, and the children of $p$ are added to the queue using the procedure \texttt{generateChildren} (line~\ref{line:updateS}), which generates the children of a node as defined in Definition~\ref{def:children}. Finally, $Res$ is returned (line~\ref{line:RprocReturn}).


 \paragraph*{\textsc{IterTD} algorithm (baseline)} Given a dataset $D$ and a ranking algorithm $R$, a size threshold $\tau_s$, a range $[k_{min}, k_{max}]$ and lower bounds $L_k$ for each $k_{min}\leqslant k \leqslant k_{max}$, a simple solution for detecting groups with biased representation based on the global representation  bounds definition utilizes Algorithm~\ref{alg:top-down}, to apply a top-down search for each $k_{min}\leq k \leq k_{max}$. Then, in each iteration report the patterns with low representation in the top-$k$ ranked items. Similarly, Algorithm~\ref{alg:top-down} can be used for the case of proportional representation, with some slight modifications (shown as comments in Algorithm~\ref{alg:top-down}).
 The objective is to report the patterns $p$ with adequate size but insufficient representation in the top-$k$ tuples $R^k(D)$, where the representation in $R^k(D)$ should be proportional to the representation in $D$. In this case, the bounds $L_k$ are not given as input, instead, a  bound for each pattern is computed based on its size in the data and a value $\alpha$. Note that the pattern's size is computed in line~\ref{line:ifSize}, and given $\alpha$ we can replace the condition in line~\ref{line:ifUnfair} with the condition $top$-$k$\_$c < \alpha\cdot s_D(p)\frac{k}{|D|}$.


We next propose more efficient algorithms for the problems.

\subsection{Global Representation Bounds}\label{sec:rankingBounds}

The key observation is that when the lower bound remains the same for $k$ and $k+1$\footnote{We assume $L_k\leqslant L_{k+1} \forall k~~ k_{min}\leqslant k < k_{max}$. This is a reasonable assumption since as $k$ increases, so is the number of tuples in the top-$k$, thus it is only logical the bounds increase as well.}, the search spaces for patterns with biased representation in the ranking result of $R$ for $k$ and $k+1$ are typically very similar. This is because the set of top-$k$ and top-$(k+1)$ tuples differ by a single tuple. Namely, increasing $k$ implies only local modifications to the search space. 
Let $T_k$ be the search tree generated to find the set of unfairly treated patterns in the top-$k$ tuples, and $R(D)[k+1]$, the $(k+1)$ element in the result of ranking $D$ using $R$. We can bound the number of nodes in $T_k$ whose size is affected by increasing~$k$. 


\begin{proposition}\label{prop:affectedNodes}
 $R(D)[k+1]$ can satisfy at most $\frac{|T_k|}{2}$ patterns (nodes) in $T_k$, where $|T_k|$ denotes the number of nodes in $T_k$.
\end{proposition}

\notFullVerOnly{The proof idea is that whenever a pattern $p$ is generated during the search, at least one pattern $p'$ with the same set of attributes ($Attr(p) = Attr(p')$) that differs from $p$ in the value assignment of a single attribute is generated as well.}

\fullVerOnly{
\begin{proof}(Sketch)
The basic idea of the proof is that whenever a pattern $p$ is generated during the search, at least one pattern $p'$ with the same set of attributes ($Attr(p) = Attr(p')$) that differs from $p$ in the value assignment of a single attribute is generated as well. This is true under the assumption that every attribute has at least $2$ values. For instance, the patterns \{Gender = M, School = MS\} and \{Gender = M, School = GP\} are both children of the pattern \{Gender = M\} and are generated when the procedure \texttt{generateChilder}(\{Gender = M\}) is invoked. Let $A_i$ be the attribute such that $A_i\in Attr(p)$ and $\{A_i = a_i\}\subseteq p$ while $\{A_i = a'_i\}\subseteq p'$. $R(D)[k+1]$ can satisfy at most one of the patterns, as the value of $A_i$ in $R(D)[k+1]$ is either $a_i$ or $a'_i$ (or possibly other value, if $|Dom(A_i)| > 2$). Since this is true for every node generated during the search, $R(D)[k+1]$ can satisfy at most $\frac{|(T_k)|}{2}$ patterns (nodes) in $T_k$.
\end{proof}
}

In that case, by starting the search for $k+1$ from the endpoint of the search for $k$, we significantly reduce the search space (and as a result, the runtime, see Section~\ref{sec:results}).

\begin{algorithm}
\footnotesize
	\DontPrintSemicolon
	\SetKwInOut{Input}{input}\SetKwInOut{Output}{output}
	\LinesNumbered
	\Input{A dataset $D$, a ranking algorithm $R$, a size threshold $\tau_s$, a range $[k_{min}, k_{max}]$ and lower bounds $L_k$ for each $k_{min}\leqslant k \leqslant k_{max}$}
	\Output{$Res$ s.t. for each $k_{min}\leqslant k \leqslant k_{max}$ $Res[k] = \{p_1, \ldots, p_n\}$ where $\forall p_i\in Res[k]$ $s_D(p_i) \geq \tau_s$ and $p_i$ is a most general pattern with  $s_{R^k(D)}(p) < L_k$} \BlankLine

	\SetKwFunction{patternCount}{\texttt{patternSize}}
	\SetKwFunction{generateChildren}{\texttt{generateChildren}}
	\SetKwFunction{parent}{\texttt{parent}}
	\SetKwFunction{matching}{\texttt{matching}}
	\SetKwFunction{searchFromNode}{\texttt{searchFromNode}}
	\SetKwFunction{update}{\texttt{update}}
	\SetKwFunction{algo}{algo}\SetKwFunction{proc}{TopDownSearch}
	\SetKwProg{myproc}{Procedure}{}{}

    $Res \gets \emptyset$  \label{line:RinitRes}\\
    
$Res[k_{min}], DRes \gets$\newline\proc{$D, R, \tau_s, k_{min}, L_{k_{min}}$} \label{line:RupdateRes1}\\
	\For{$k=k_{min}+1$ to $k_{max}$\label{line:RforKStart}}
	{
	    \If{$L_{k-1} < L_{k}$\label{line:Rboundchanges}}
	    {
	        $Res[k], DRes \gets $\newline\proc{$D, R^k(D), \tau_s, L_{k}$}
	    }
	   \Else
	   {\label{line:rbounddoesnotchange}
	   $Res[k]\gets Res[k-1]$\label{line:RupdateRes2}\\
            \ForEach{$b\in \{p\in Res[k-1] \mid  R(D)[k] \text{ satisfies } p\} \cup DRes$ \label{line:RforKstart}} 
            {
                 $Res[k], DRes \gets $\newline\searchFromNode($b, Res[k], DRes$) \label{line:RcheckB}
            }
        }
	}
	\Return $Res$ \label{line:Rreturn}\\
	\caption{\textsc{GlobalBounds}. Detecting groups with biased representation based on global bounds}\label{alg:ranking}
\end{algorithm}

\paragraph*{\textsc{GlobalBounds} algorithm} Our optimized algorithm for the problem of detecting groups with global representation bias in the top-$k$, \textsc{GlobalBounds}, is depicted in  Algorithm~\ref{alg:ranking}. \textsc{GlobalBounds} 
starts by initializing the result set map $Res$ (line~\ref{line:RinitRes}). It then performs a top-down search for the case where $k=k_{min}$ (line~\ref{line:RupdateRes1}). This is done using the procedure \texttt{TopDownSearch}, similar to the algorithm depicted in Algorithm~\ref{alg:top-down}, with a minor addition.  
It maintains a set $DRes$ of patterns $p$ reached during the search with size below the lower bound in top-$k$, that are not part of the result set since it already contains an ancestor of $p$.  \texttt{TopDownSearch} returns both, the result set of the search $Res$, and the set $DRes$.
When $k$ increases (and $L_k$ is kept intact), the algorithm will utilize this set to initiate a local search in the pattern graph. 

Next, the algorithm preforms the search for each $k$ from $k=k_{min}+1$ through $k=k_{max}$ (lines~\ref{line:RforKStart}--\ref{line:RcheckB}).
For each $k$, if the bound increases,  \texttt{TopDownSearch} is used to perform a new top-down search. Otherwise, The algorithm considers only patterns from $DRes$ and patterns from $Res[k-1]$ that the newly inserted tuple $R(D)[k]$ satisfies (line~\ref{line:RforKstart}). This is because only their sizes in the top-$k$ are affected by the new tuple (at most half of the tree, based on Proposition~\ref{prop:affectedNodes}). For each such pattern, the algorithm applies the procedure \texttt{searchFromNode}  (line~\ref{line:RcheckB}) to resume the search in the relevant parts of the graph. This search updates $Res[k]$ and $DRes$. Finally, $Res$ is returned (line~\ref{line:Rreturn}).

\begin{proposition}\label{prop:global_correctness}
\textsc{GlobalBounds} returns the set of all most general patterns $p$ with bias representation using global bounds in the top-$k$ for each $k$ in the given range.
\end{proposition}

The proof is by induction on $k$ with a base case of $k=k_{min}$. Details are omitted due to space constraints.
\begin{example}
    Consider again $D$ and $R$ from the running example.
    Assume we are given the size threshold $\tau_s = 4, k_{min} = 4, k_{max} = 5$, and the lower bounds $L_4 = L_5 = 2$. At the end of the top-down search for $k=4$, the result set $Res[4]$ contains (among others) the patterns \{Address = U\} and \{Failures = 1\}, that appears only once in the top-$4$ tuples (namely, below the lower bound). 
    $DRes$ contains, for instance, the patterns \{Gender = F, Address = U\}, \{Gender = M, Address = U\}, \{Gender = F, Failures = 1\} and \{Address = R, Failures = 1\}. These patterns were generated during the top-down search and have ancestors in $Res[4]$ (\{Address = U\} and \{Failures = 1\}). Next, the algorithm turns to compute patterns with biased representation for $k=5$. The new tuple in the top-$5$ is tuple $14$. It matches the patterns \{Address = U\} and \{Failures = 1\} in $Res[4]$. Thus the algorithm performs the search starting from those nodes. Their sizes in the top-$5$ exceed the lower bound. In this search, these two patterns are extracted from the result set and the pattern \{Address = U, Failures = 1\} is added.
    From the set $DRes$, the patterns
    \{Gender = F, Address = U\}, \{Gender = M, Address = U\}, \{Gender = F, Failures = 1\} and \{Address = R, Failures = 1\} are added to the result set $Res[5]$, as their sizes in the top-$5$ tuples are still below the threshold $L_5$ and their respective ancestors are removed from the result set.
\end{example}

\subsection{Proportional Representation}\label{sec:proportional}

We next consider the problem of detecting groups with biased proportional representation 
as depicted in Problem~\ref{problem:ranking:proportional}. The inputs are a dataset $D$, a ranking algorithm $R$, a range $[k_{min}, k_{max}]$, a size threshold $\tau_s$ and $\alpha\in \mathbb{R}$. 
  The objective is to report the patterns $p$ with adequate size in $D$, but insufficient representation in the top-$k$ tuples $R^k(D)$, where the representation in $R^k(D)$ should be proportional to the representation of $p$ in $D$. 


First, note that the optimized solution presented for the case of global representation bounds depicted in Section~\ref{sec:rankingBounds} is not applicable in this case. Recall that \textsc{GlobalBounds} (Algorithm~\ref{alg:ranking}) aims at reducing the search space by avoiding searching areas in the pattern graph that were not changed between consecutive iterations. When the bound remains unchanged ($L_k=L_{k+1}$), patterns that the ($k+1$) tuple in the ranking does not satisfy, are not affected, and can be eliminated from the search. This is not the case for proportional representation, as the bound for each pattern depends on $k$ as well.



Recall that the goal is to find patterns $p$ such that $s_{R^k(D)}(p) < \alpha\cdot s_D(p)\cdot\frac{k}{|D|}$, and note that $\alpha$ and $s_D(p)$ do not change during the computations. Thus, the inequality holds depending on the value of $k$. 
Given $R$, $D$, $\alpha$, a pattern $p$, and a value $k$ such that $s_{R^k(D)}(p) \geqslant \alpha\cdot s_D(p)\cdot\frac{k}{|D|}$, we denote by $\minK$ the minimal value for $k$ such that the inequality does not hold when fixing the value of $s_{R^k(D)}(p)$. Namely, the minimal value such that $s_{R^k(D)}(p) < \alpha\cdot s_D(p)\cdot\frac{\minK}{|D|}$.

\begin{example}
    Let $\alpha=0.9$. For $D$ and $R$ from our running example, $p=$\{Gender = F\} satisfies the inequality for $k=4$ since $s_{R^k(D)}(p) = 2 > 1.8 = 0.9\cdot 8\cdot\frac{4}{16}$. $\minK=5$ in this case since $0.9\cdot 8\cdot\frac{5}{16} = 2.25$.
\end{example}



Intuitively, if $k$ is increased up to $\minK$ but the number of tuples satisfying $p$ remains the same, then the representation of $p$ is biased in the ranking result. If there is no ancestor of $p$ in the result set, then $p$ should be added to it.
To this end, our optimized algorithm, \textsc{PropBounds}, computes for each pattern in the search tree its corresponding $\minK$ value. It maintains a set $\mathcal{K}$ which indicates patterns that potentially, if they do not satisfy the $\minK$ element in the ranking, should be added to the result set. $\mathcal{K}$ contains patterns in a branch of the search tree whose $\minK$ values are monotonically decreasing. 
A pattern $p$ in $\mathcal{K}$ with the corresponding $\minK$ value should be extracted from $\mathcal{K}$ and added to the result set when the computation reaches $k = \minK$ if $s_{R^k(D)}(p) = s_{R^{k-1}(D)}(p)$ (i.e., $p$ does not satisfy $R(D)[k]$). 

\begin{algorithm}
\small
	\DontPrintSemicolon
	\SetKwInOut{Input}{input}\SetKwInOut{Output}{output}
	\LinesNumbered
	\Input{A dataset $D$, a ranking algorithm $R$, a size threshold $\tau_s$, a range $[k_{min}, k_{max}]$ and $\alpha\in \mathbb{R}$}
	\Output{$Res$ s.t. for each $k_{min}\leqslant k \leqslant k_{max}$ $Res[k] = \{p_1, \ldots, p_n\}$ where $\forall p_i\in Res[k]$ $s_D(p_i) \geq \tau_s$ and $p_i$ is a most general pattern with  $s_{R^k(D)}(p) <\alpha\cdot s_D(p)\frac{k}{|D|}$} \BlankLine

	\SetKwFunction{patternCount}{\texttt{patternSize}}
	\SetKwFunction{generateChildren}{\texttt{generateChildren}}
	\SetKwFunction{parent}{\texttt{parent}}
	\SetKwFunction{matching}{\texttt{matching}}
	\SetKwFunction{searchFromNode}{\texttt{searchFromNode}}
	\SetKwFunction{update}{\texttt{update}}
	\SetKwFunction{algo}{algo}\SetKwFunction{proc}{TopDownSearch}
	\SetKwProg{myproc}{Procedure}{}{}
	\SetKwFunction{selectiveTD}{\texttt{selectiveTD}}

    $Res \gets \emptyset$  \label{line:prop_initRes}\\

$Res[k_{min}], \mathcal{K}, DRes \gets$\newline\proc{$D, R, \tau_s, k_{min}, \alpha$} \label{line:prop_first_search}\\
	\For{$k=k_{min}+1$ to $k_{max}$\label{line:prop_forKStart}}
	{
	    $Res[k]\gets Res[k-1]$ \label{line:prop_prevRes}\\
     
	    $Res[k], \mathcal{K}, DRes \gets$\newline\selectiveTD($D, R, \tau_s, k, \alpha, R(D)[k], \mathcal{K}, DRes$)\label{line:prop_search_selective}\\
     \ForEach{$p \in DRes \bigcup_{\mathcal{K}[p']=k} \{p'\}$ such that $R(D)[k]$ doesn't satisfy $p$  \label{line:prop_forEach}}
     {
        \update($Res, p$) \label{line:prop_updateRes}
     }
	}
	\Return $Res$ \label{line:prop_return}\\
	\caption{\textsc{PropBounds}. Detecting groups with biased proportional representation}\label{alg:prop}
\end{algorithm}

\paragraph*{\textsc{PropBounds} algorithm}
\textsc{PropBounds} (Algorithm~\ref{alg:prop}) operates as follows. Similarly to \textsc{GlobalBounds}, it 
starts by initializing the result set map $Res$ (line~\ref{line:prop_initRes}) and applying a top-down search for $k_{min}$ (line~\ref{line:prop_first_search}), as depicted in procedure \texttt{TopDownSearch} (with the required modification for proportional bounds), but in addition to sets $Res, DRes$, it also maintains the set $\mathcal{K}$. Then, the algorithm iterates over the values of $k$ from $k_{min}+1$ to $k_{max}$ (lines~\ref{line:prop_forKStart}~--~\ref{line:prop_updateRes}).
In each iteration, it first initializes $Res[k]$ with the result from the previous iteration $Res[k-1]$ (line~\ref{line:prop_prevRes}). Then the algorithm applies a (partial) search from the root using the procedure \texttt{selectiveTD} (line~\ref{line:prop_search_selective}). This search ignores the areas in the tree that are not affected by $R(D)[k]$.
The result of the procedure is used to update the result set, $\mathcal{K}$, and $DRes$.
The algorithm then iterates over the patterns in $DRes$ (patterns reached during the search that has an ancestor in the result) 
and all patterns in $\mathcal{K}$ with $\minK = k$ that are not affected by $R(D)[k]$, to determine the changes to the result set (lines~\ref{line:prop_forEach}~--~\ref{line:prop_updateRes}).
Finally, $Res$ is returned (line~\ref{line:prop_return}).




\begin{proposition}
\textsc{PropBounds} returns the set of all most general patterns $p$ with bias proportional representation in the top-$k$ for each $k$ in the given range.
\end{proposition}
Similarly to Proposition~\ref{prop:global_correctness}, the proof is by induction on $k$ with a base case of $k=k_{min}$. Due to space constraints, the details are omitted.

\begin{example}
    Consider again $D$ and $R$ from the running example. 
    Assume we are given the size threshold $\tau_s = 5, k_{min} = 4, k_{max} = 5$, and $\alpha = 0.9$. At the end of the top-down search for $k=4$, the result set $Res[4]$ consists of the patterns \{School = GP\}, \{Address = U\} and \{Failures = 1\}. For each one of them, $s_D(p) = 8$, and thus the bound on $s_{R^k(D)}(p)$ is $\alpha\cdot  s_D(p)\cdot\frac{k}{|D|} = 0.9\cdot 8\cdot\frac{4}{16} = 1.8$, but $s_{R^k(D)}(p)$ is only $1$. The set $\mathcal{K}$ consists of \{Gender = M\} and \{Gender = F\}, both with $\minK$ of $5$, and \{School = MS\} and \{Address = R\} with $\minK = 7$. Note that the pattern \{School = MS, Address = R\} was generated in the first top-down search, but was not added to $\mathcal{K}$ since its $\minK$ value is 9, higher than its parent in the search tree \{School = MS\}.
    
    When $k$ is increased to $5$, the algorithm reexamines only the patterns \{Gender = M\},  \{School = MS\}, \{Address = U\} and \{Failures = 1\}  that are affected by the $R(D)[5]$ (tuple $14$). The patterns \{Address = U\} and \{Failures = 1\} remain in the result set for $k=5$ even though their size in the top-$5$ is larger, since the bound for $k=5$ increases as well. Finally, the pattern \{Gender = F\} is added to $Res[5]$ based on the information from $\mathcal{K}$ (it is stored in $\mathcal{K}$ with $\minK = 5$).
\end{example}

%% file: explanations.tex
\section{Result Analysis}\label{sec:explanations}

With the results of our algorithm in hand, an analyst may wish to understand the cause of the bias in the representation of the detected groups. We propose a method to provide such explanations utilizing the notion of Shapley values~\cite{shapley20207}.
Shapley value is a concept adopted from game theory to explain the effect of different attributes on the output of a model for a given input. The use of Shapley values has recently gained popularity in the field of interpretability and explainability of ML models~\cite{StrumbeljK14,LundbergL17}. Given a regression model (or a classifier with probabilities) $M$, Shapley values are used to evaluate the contribution of each attribute on the output of $M$ for a given input $t$. This is done by computing the weighted marginal contribution of each attribute value using all possible subsets of attribute values.

Intuitively, an explanation for the bias may be the values that affected the ranking of tuples in the given group. To this end, we propose a method for explanations that consists of two parts: the first identifies the attributes with the highest effect on the ranking of tuples in the given group (using Shapley values), and then we compare the values distribution of these attributes in the top-$k$ and the biased represented group. In order to adopt the use of Shapley values, we need to tackle two challenges. The first is to adjust our problem's setting, where we are given a ranking algorithm $R$ (as a black box) rather than a regression model. Second, Shapley values are used to explain the contribution of the attribute values of a single tuple, whereas we are interested in explaining the (inadequate) representation of a group of tuples (in the top-$k$).

To address the first challenge we compute a regression model $M_R$ that simulates the process of $R$ and can be used to approximate the effect of attribute values of a given tuple $t$ on $t$'s ranking by computing the Shapley values of $M_R(t)$. To this end we define $D_R = \{(t, R(D)[t])\mid t\in D\}$, where $R(D)[t]$ is the ranking of $t$ in $R(D)$, and use it to train a regression model $M_R$. Then, given a pattern $p$ such that $p$ was returned by one of our algorithms for detecting groups with biased representation for a given $k$, to explain the result, we compute the Shapley values $(s^t_1,\ldots,s^t_m)$ for each tuple $t$ such that $t$ satisfies $p$, namely, for each tuple in the detected group. We then aggregate the results into a single Shapley value vector $(s_1,\ldots,s_m)$ for the pattern $p$ such that 
\vspace{-2mm}
$$
s_i = \frac{\sum_{t \text{ s.t. } t \text{ satisfies } p}s^t_i}{s_{D}(p)}
$$
\vspace{-2mm}

To show the differences between the pattern $p$ and the top-$k$ patterns, we visualize the value distribution of attributes with large Shapley values of tuples that satisfy the pattern $p$ compared to their distribution among the tuples in the top-$k$.
In Section~\ref{sec:shapley_exp} we show that using our method 
we are able to disclose the attributes that were used for ranking (and thus affect the representation of groups in the top-$k$) when the ranking model is given as a black box. Moreover, we show that the value distribution in the attribute identified as most significant in the ranking is different for groups detected by our algorithms than for the top-$k$ tuples, which indicates the identified attributes values explain the results.

%% file: exp.tex
\section{Experiments}\label{sec:exp}

We experimentally examine the proposed solutions using three real-life datasets.
We start with our setup and then present a quantitative experimental study whose goal is to assess the scalability of our algorithms. In particular, we examine our algorithms' performance for each fairness definition as a function of the number of attributes, pattern's size threshold, and range of $k$. We then demonstrate our proposed method for the analysis of the results presented in Section~\ref{sec:explanations}. We conclude with a comparison to the framework presented in~\cite{pastor2021identifying}, showing the differences in the results between our algorithms and the algorithm of~\cite{pastor2021identifying} through a case study.

\subsection{Experiment Setup}
\paragraph{Datasets} We used three real datasets with different numbers of tuples and attributes as follows. 

\begin{itemize}[leftmargin=1em,labelwidth=*,align=left]
\item 
\textbf{The COMPAS Dataset}\footnote{\url{https://www.propublica.org/datastore/dataset/compas-recidivism-risk-score-data-and-analysis}} was collected and published by ProPublica as part of their investigation into racial bias in criminal risk assessment software. 
It contains the demographics, recidivism scores produced by the COMPAS software, and criminal offense information for 6,889 individuals. We used up to 16 attributes eliminating attributes such as names, ids, dates, etc.

\item 
\textbf{Student Performance Dataset (Student dataset)}\footnote{\url{https://archive.ics.uci.edu/ml/datasets/student+performance}} shows the performance of students in secondary education of two Portuguese schools as described in Example \ref{ex:running}. We considered in the experiment the data fragment with information regarding the Math exam
(395 tuples and 33 attributes).

\item 
\textbf{German Credit Dataset}\footnote{\url{https://archive.ics.uci.edu/ml/datasets/Statlog+(German+Credit+Data)}} with financial and demographic information about 1,000 loan applicants with 20 attributes. 
It was originally used in the context of classification, where each application is classified as a good or bad credit risk.


\end{itemize}

\paragraph*{Compared algorithms} 
We evaluate the performance, in terms of the running time of our proposed algorithms.
\begin{itemize}[leftmargin=1em,labelwidth=*,align=left]
\item \textbf{IterTD (baseline).} The simple solution for detection of groups with biased representation, which iteratively applies a top-down search as depicted in Section~\ref{sec:top-down}.
\item \textbf{\textsc{GlobalBounds} (Algorithm \ref{alg:ranking}).} The algorithm for detecting groups with biased representation based on global bounds as described in Section~\ref{sec:rankingBounds}. 
\item \textbf{\textsc{PropBounds} (Algorithm \ref{alg:prop}).} The algorithm for detecting groups with biased representation  based on proportional representation bounds as described in Section~\ref{sec:proportional}. 
\end{itemize}


\paragraph*{Parameters setting} 

For space constraints, unless stated otherwise, we report the result for the following set of default parameters: $\tau_s=50$, $k_{min}=10$, $k_{max}=49$, and the lower bounds are $10$ for $10\leq k < 20$, $20$ for $20\leq k< 30$, $30$ for $30 \leq k< 40$ and $40$ for $40 \leq k< 50$ for global bounds, and $\alpha=0.8$ for proportional representation. 
\rev{The reported results reflect the algorithm's performance under expected and typical usage scenarios. Following the goals of fairness in ranking (ensuring fairness for any position), we set gradually increasing bounds on group representation in the top-$k$ ranked items as $k$ increases. Since we aim at reporting the detected groups to the user, we set the parameters such that the number of reported groups in most cases is between $1$ to $100$.} The number of attributes was set to be the maximal number the baseline solution could handle  \rev{and continuous attributes, e.g., age, were bucketized equally into $3~-~4$ bins, based on their domain and values. 
We note that the selection of bucketization affects the patterns graph size and may also affects the possible group definitions and their representation in the top-$k$, which could also affect the running time. However, in this work, we assume that the attribute values used for group definitions are categorical (i.e., the bucketization is given).
 We experimentally evaluated the algorithms using different parameter settings and observed similar trends.}

\paragraph*{Ranking Algorithms}
The Student dataset was ranked based on the value of the attribute \textsf{G3} showing the  student's math final grades.
For the COMPAS dataset, we performed a similar ranking method as in~\cite{asudeh2019designing}:
We normalized attribute values \textsf{c\_days\_from\_compas}, \textsf{juv\_other\_count}, \textsf{days\_b\_screening\_arrest}, \textsf{start}, \textsf{end}, \textsf{age}, and \textsf{priors\_count} as scoring attributes.
Values are normalized as $(val - min)/(max - min)$.
Higher values correspond to higher scores, except for age.
Tuples are ranked descendingly according to their scores. For the German Credit dataset, we used the ranking presented in \cite{YangS17} based on creditworthiness.

All experiments were performed on a macOS machine with a 2.8 GHz Quad-Core Intel Core i7 CPU and 8GB memory. The algorithms were implemented using Python3.7.

\subsection{Experimental results}\label{sec:results}

Both \textsc{GlobalBounds} and \textsc{PropBounds} run much faster than the baseline, particularly as the number of attributes increases and the baseline becomes exponentially more expensive. Details below. 

\begin{figure*}[!htb]
	\centering
	\begin{subfigure}[t]{0.3\textwidth}
		\includegraphics[width = 0.9\linewidth]{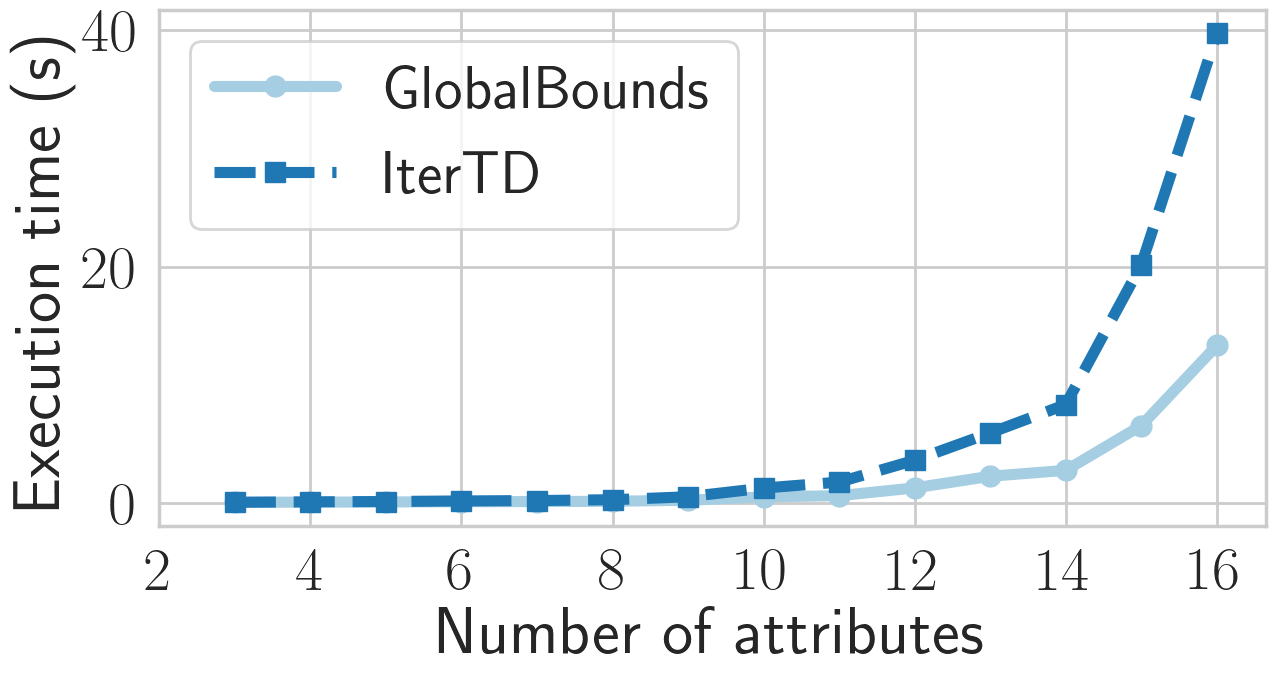}
		\caption{COMPAS dataset}
		\label{fig:compas_attr_num_ranking}
	\end{subfigure}%
	\begin{subfigure}[t]{0.3\textwidth}
		\includegraphics[width =0.9\linewidth]{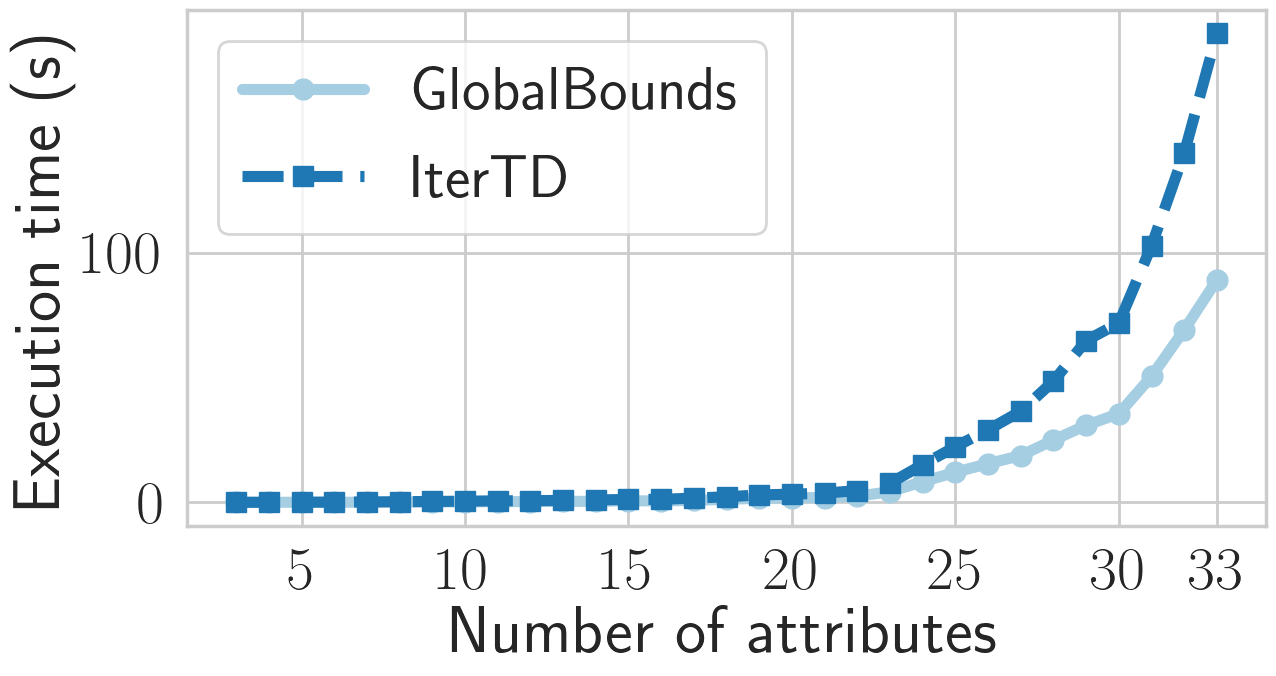}
		\caption{Students dataset}
		\label{fig:students_attr_num_ranking}
	\end{subfigure}%
	\begin{subfigure}[t]{0.3\textwidth}
		\includegraphics[width =0.9\linewidth]{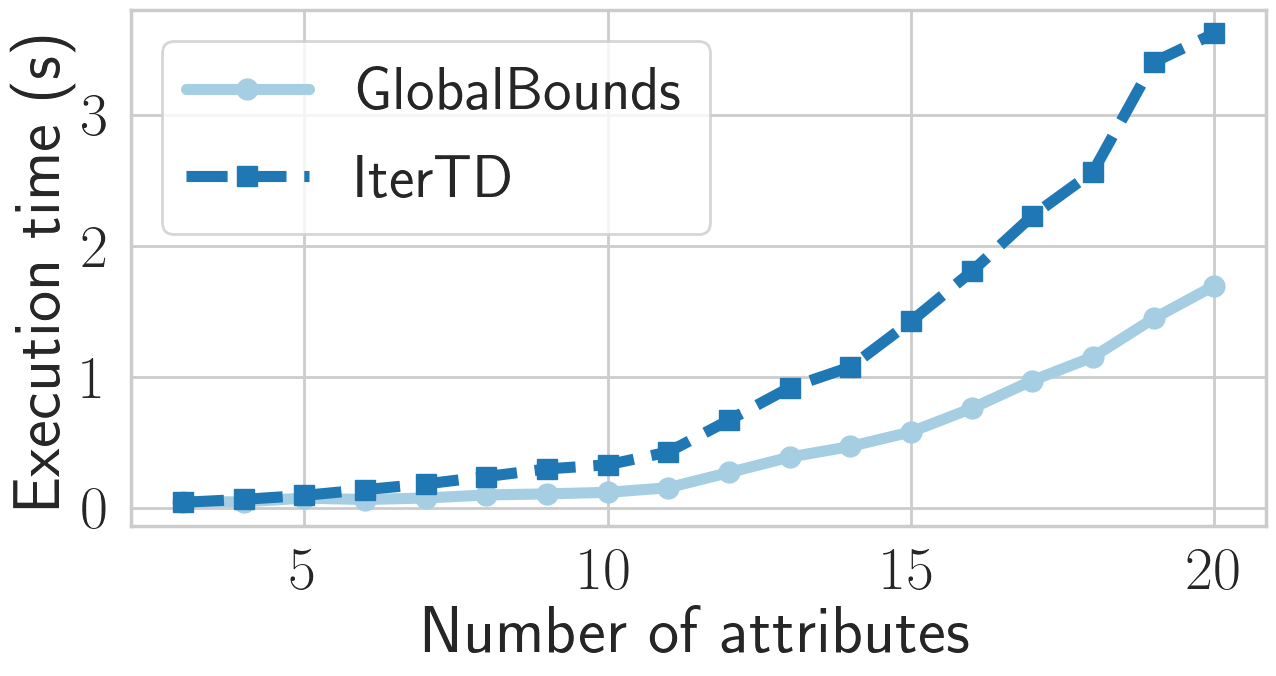}
		\caption{German Credit}
		\label{fig:german_attr_num_ranking}
	\end{subfigure}%
	\caption{Running time as a function of number of attributes - Ranking with global bounds} 
 \vspace{-1mm}
 \label{fig:attr_num_ranking}
\end{figure*}


\begin{figure*}[!htb]
	\centering
	\begin{subfigure}[t]{0.3\textwidth}
		\includegraphics[width = 0.9\linewidth]{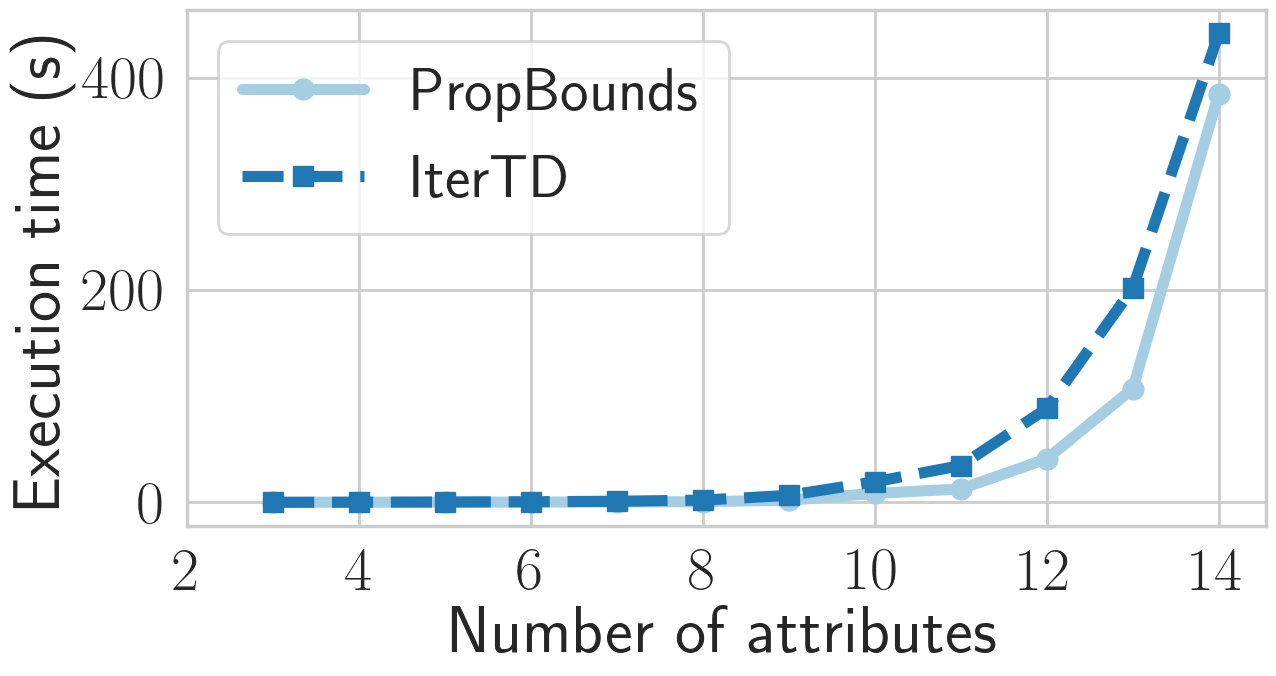}
		\caption{COMPAS dataset}
		\label{fig:compas_attr_num_ranking_proportional}
	\end{subfigure}%
	\begin{subfigure}[t]{0.3\textwidth}
		\includegraphics[width =0.9\linewidth]{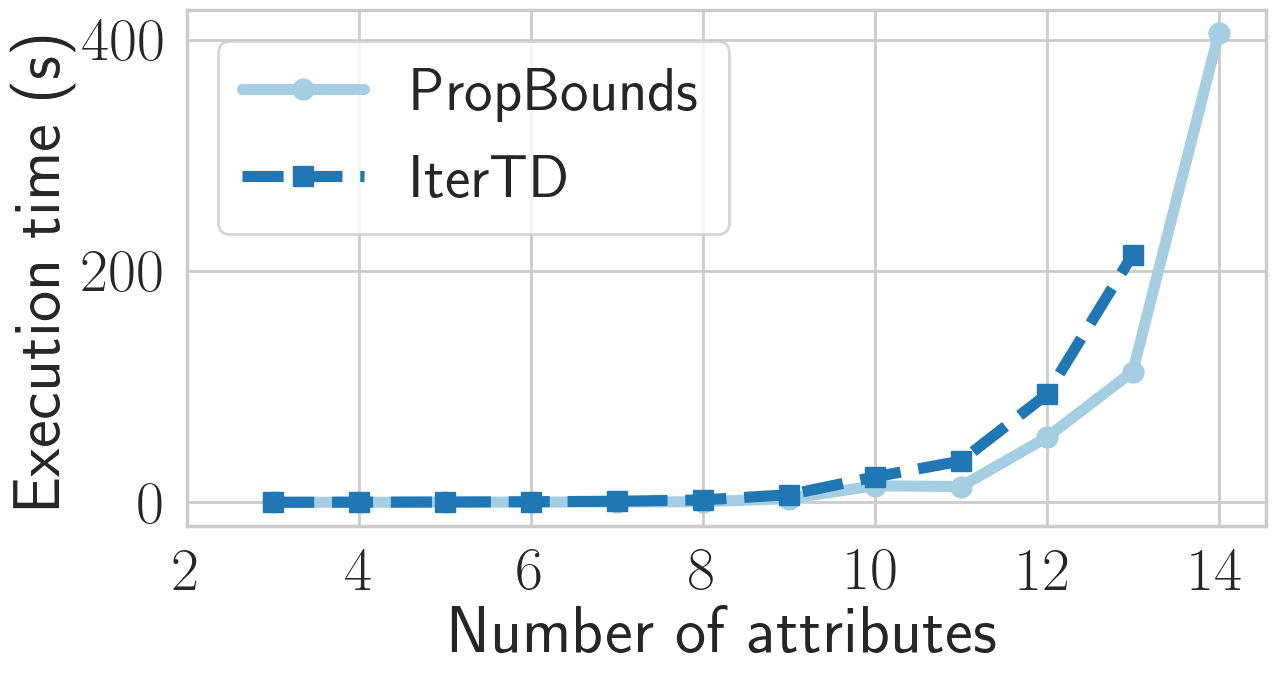}
		\caption{Student dataset}
		\label{fig:students_attr_num_ranking_proportional}
	\end{subfigure}%
		\begin{subfigure}[t]{0.3\textwidth}
		\includegraphics[width =0.9\linewidth]{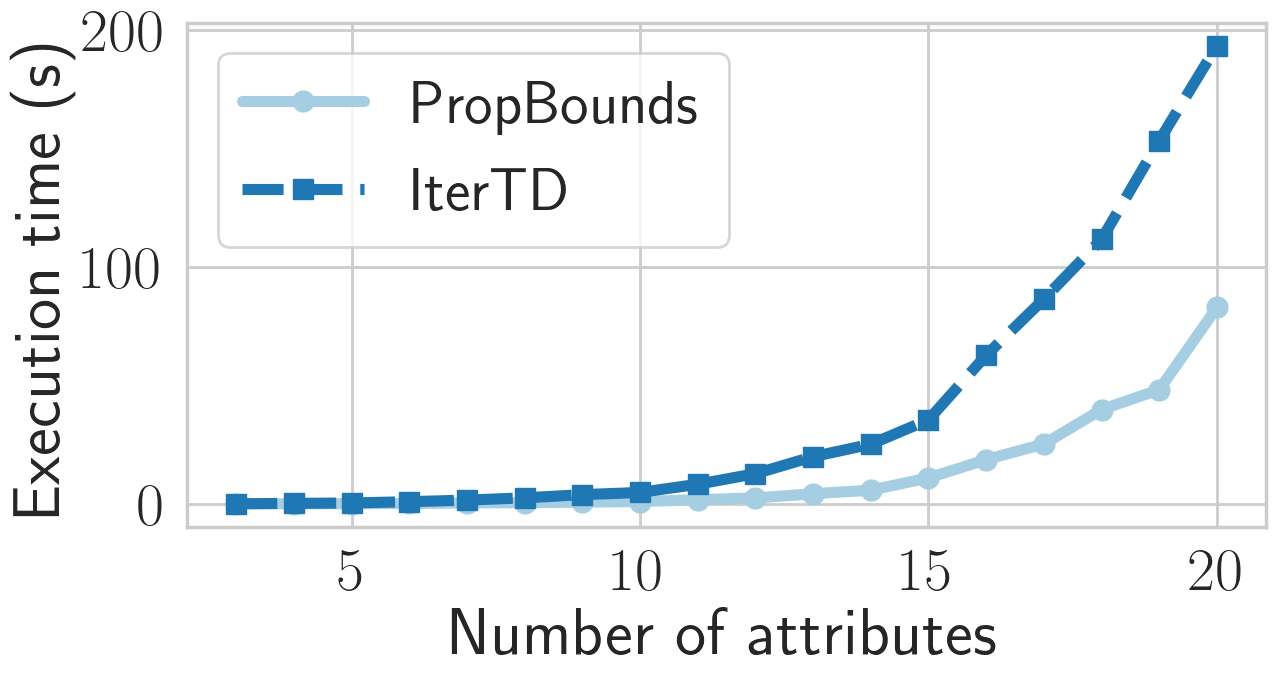}
		\caption{German Credit}
		\label{fig:german_attr_num_ranking_proportional}
	\end{subfigure}%
	\caption{Running time as a function of the number of attributes - Ranking with proportional representation} 
  \vspace{-1mm}
  \label{fig:attr_num_ranking_proportional}
\end{figure*}

\paragraph*{Number of attributes} The first set of experiments aims to study the effect of the number of attributes on the running time. To this end, we varied the number of attributes in the datasets from $3$ to $|\mathcal{A}|$ where $\mathcal{A}$ is the set of all attributes in the dataset.
The number of attributes (along with their cardinality) determines the number of possible patterns, and as a result, the size of the search space. Thus, as the number of attributes increases, we expect to see a steep growth in the running time. The results (using a 10-minute timeout) are presented in Figures~\ref{fig:attr_num_ranking}--\ref{fig:attr_num_ranking_proportional}.
Indeed, in all cases, we observed a rapid increase in the running time, while \textsc{GlobalBounds} and \textsc{PropBounds} outperform 
\textsc{IterTD}.

\begin{figure*}[!htb]
	\centering
	\begin{subfigure}[t]{0.3\textwidth}
		\includegraphics[width = 0.9\linewidth]{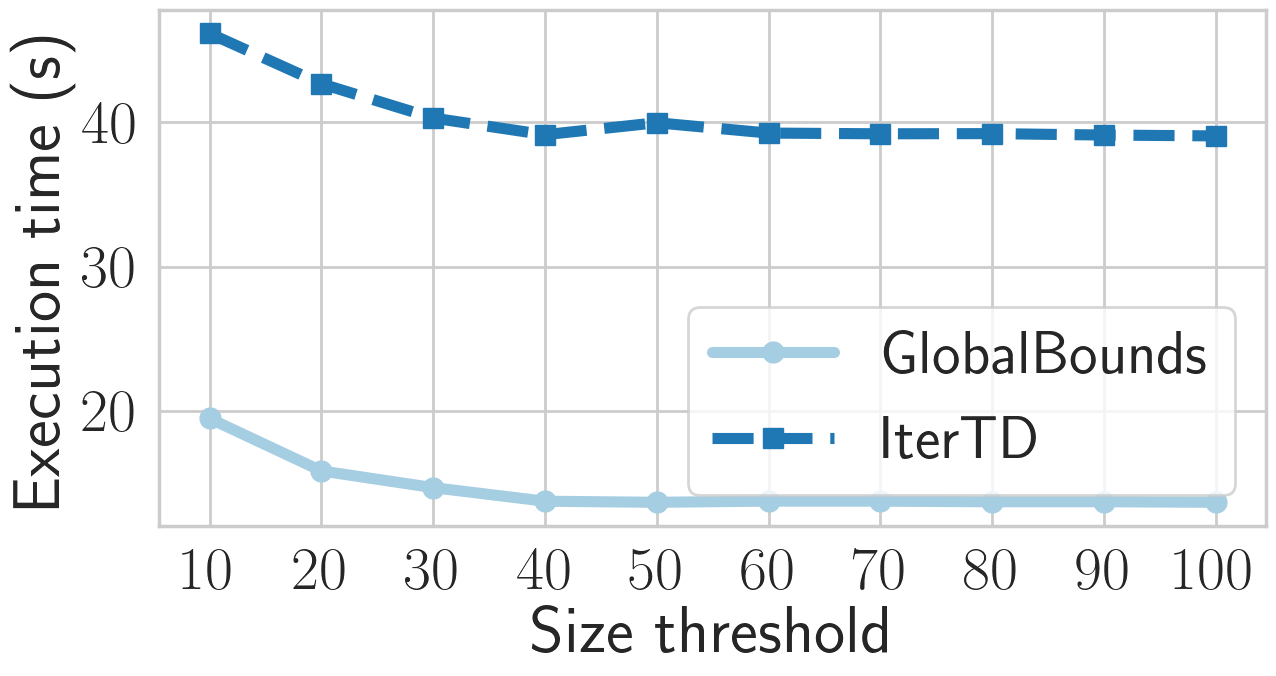}
		\caption{COMPAS dataset}
		\label{fig:compas_size_t_ranking}
	\end{subfigure}%
	\begin{subfigure}[t]{0.3\textwidth}
		\includegraphics[width =0.9\linewidth]{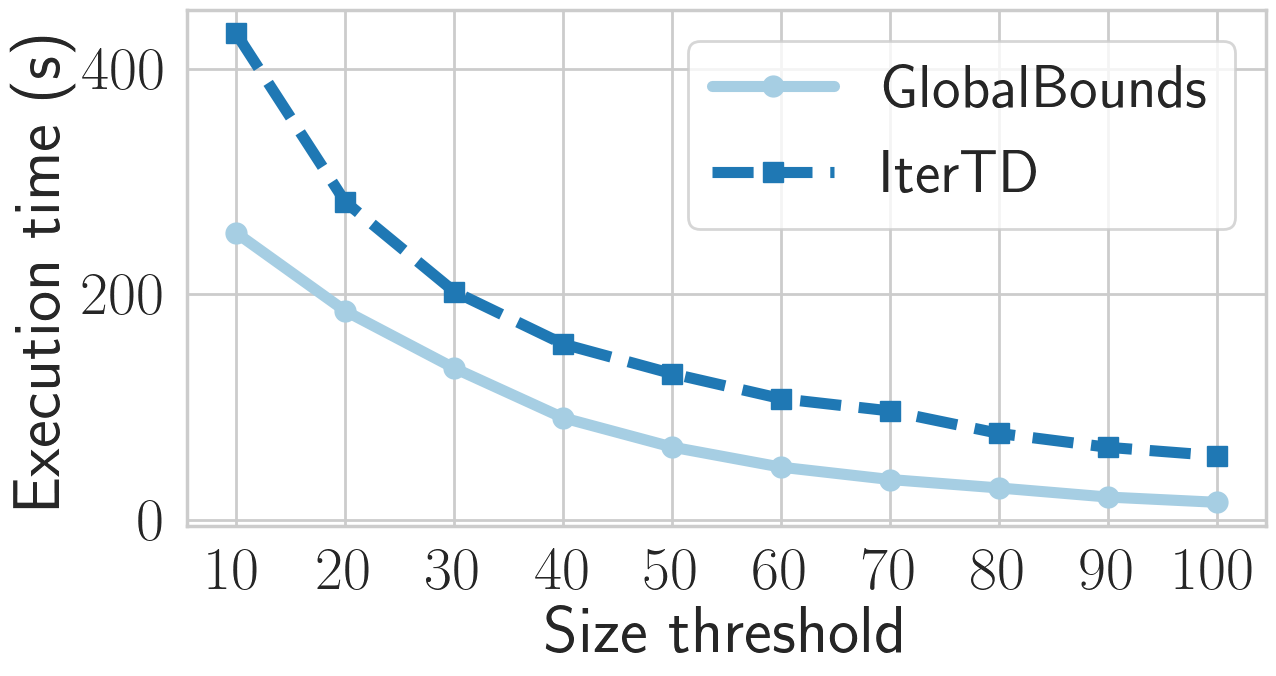}
		\caption{Students dataset}
		\label{fig:students_size_t_ranking}
	\end{subfigure}%
	\begin{subfigure}[t]{0.3\textwidth}
		\includegraphics[width =0.9\linewidth]{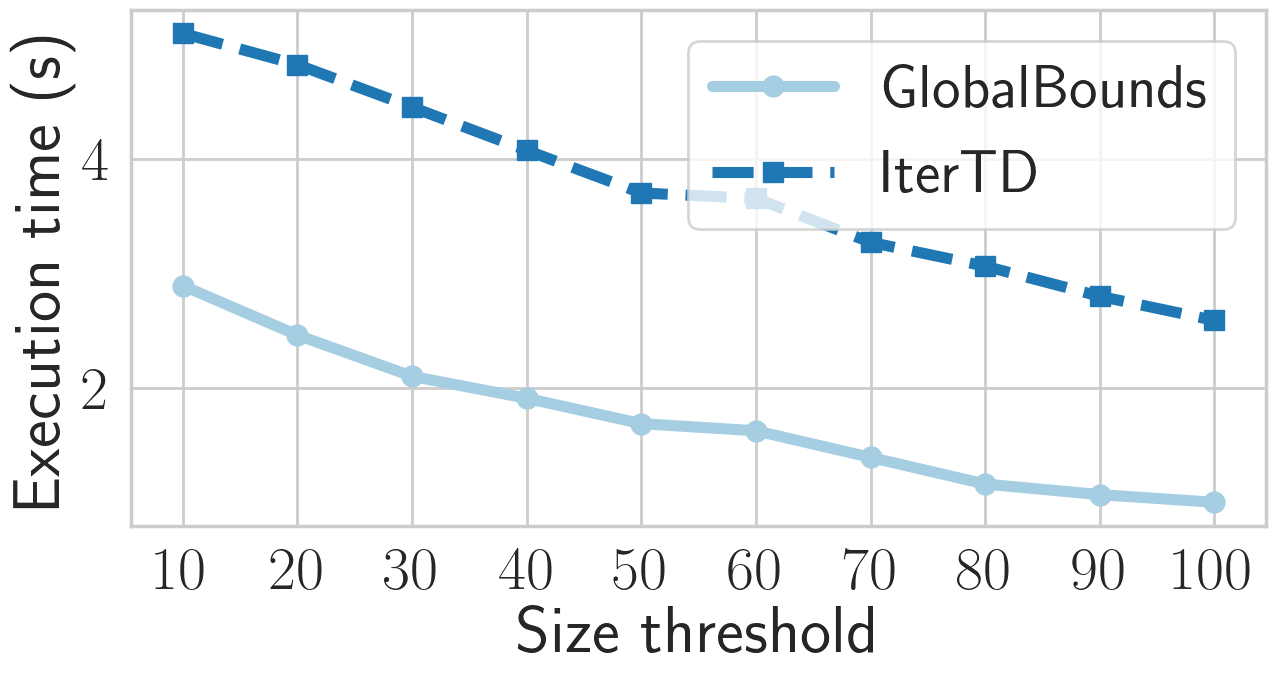}
		\caption{German Credit}
		\label{fig:german_size_t_ranking}
	\end{subfigure}%
	\caption{Running time as a function of the size threshold $\tau_s$ - Ranking with global bounds}
  \vspace{-1mm}
  \label{fig:size_t_ranking}
\end{figure*}


\begin{figure*}
	\centering
	\begin{subfigure}[t]{0.3\textwidth}
		\includegraphics[width = 0.9\linewidth]{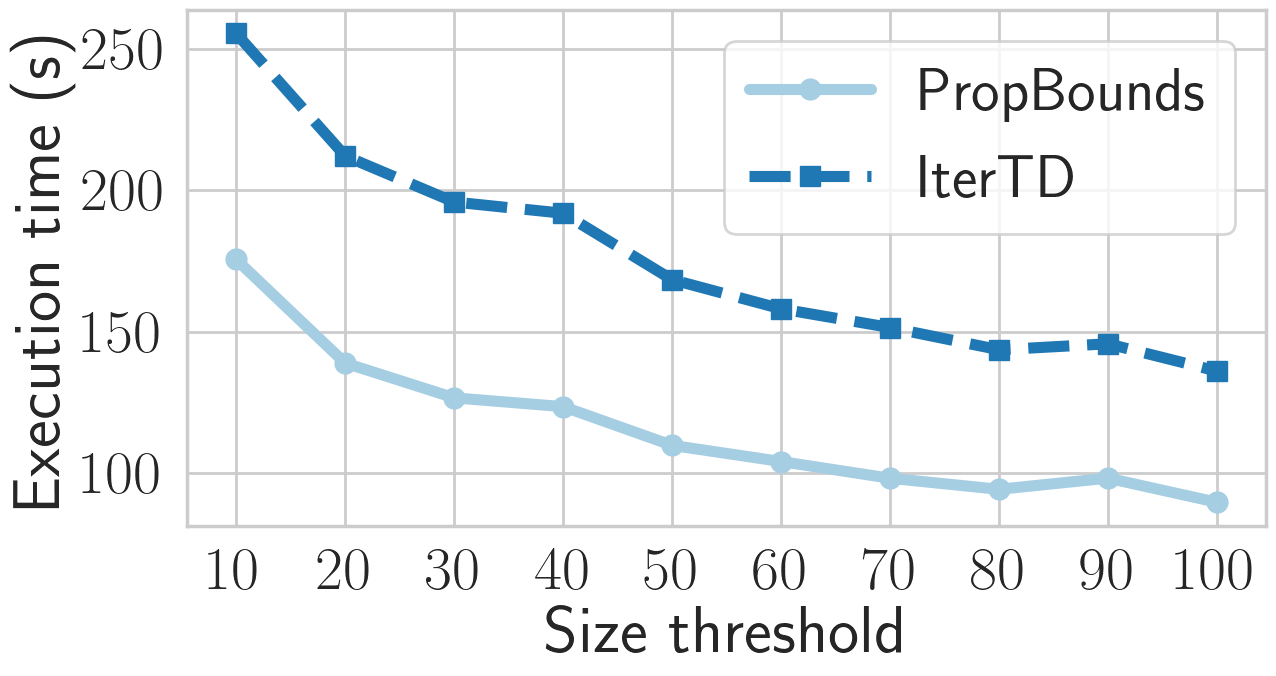}
		\caption{COMPAS dataset}
		\label{fig:compas_size_t_ranking_proportional}
	\end{subfigure}%
	\begin{subfigure}[t]{0.3\textwidth}
		\includegraphics[width =0.9\linewidth]{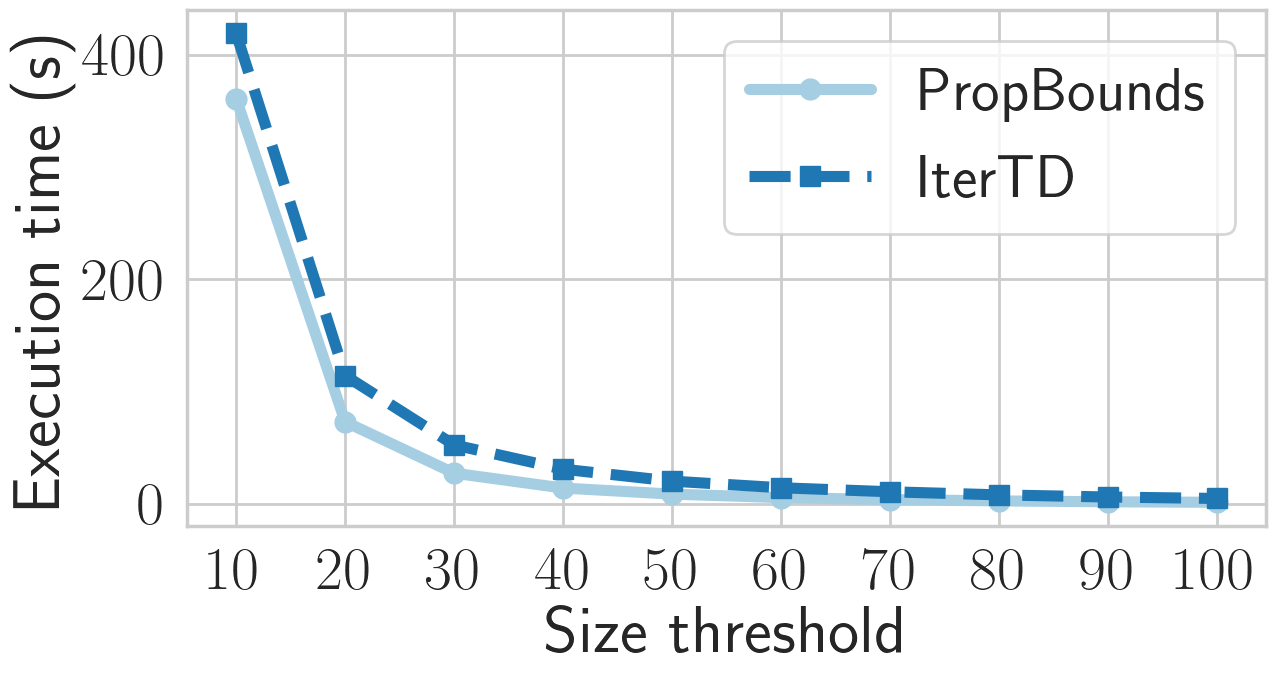}
		\caption{Students dataset}
		\label{fig:students_size_t_ranking_proportional}
	\end{subfigure}%
		\begin{subfigure}[t]{0.3\textwidth}
		\includegraphics[width =0.9\linewidth]{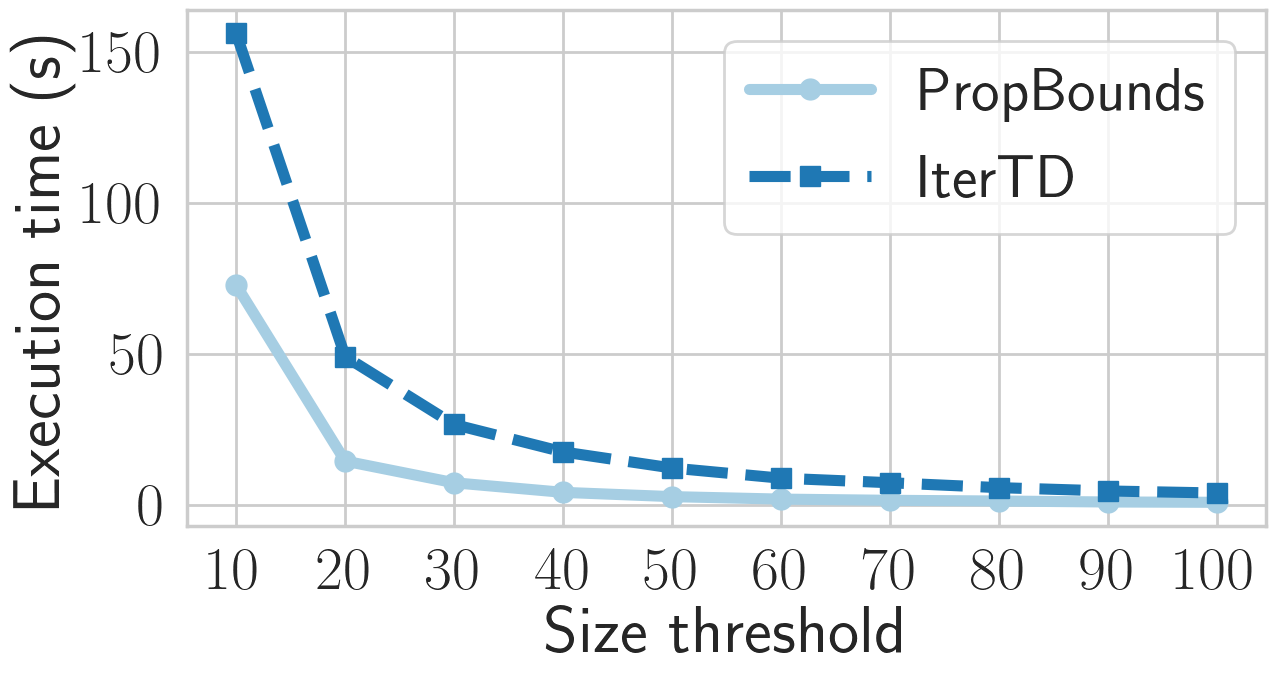}
		\caption{German Credit}
		\label{fig:german_size_t_ranking_proportional}
	\end{subfigure}%
	\caption{Running time as a function of the size threshold $\tau_s$ - Ranking with proportional representation} 
 \vspace{-2mm}
 \label{fig:size_t_ranking_proportional}
\end{figure*}

\paragraph*{Size threshold} In the next set of experiments, we assessed the effect of the size threshold $\tau_s$ on the running time. To this end, we varied the size threshold from $10$ to $100$ while using the default values for the rest of the parameters. The results are presented in Figures~\ref{fig:size_t_ranking} and \ref{fig:size_t_ranking_proportional}. We observed a decrease in the running times of the algorithms. This is because the number of patterns satisfying the size threshold decreases as the threshold increases, and as a result, the search space is decreased as well.  In all cases, \textsc{GlobalBounds} and \textsc{PropBounds} outperform 
\textsc{IterTD}.

\begin{figure*}
	\centering
	\begin{subfigure}[t]{0.3\textwidth}
		\includegraphics[width = 0.9\linewidth]{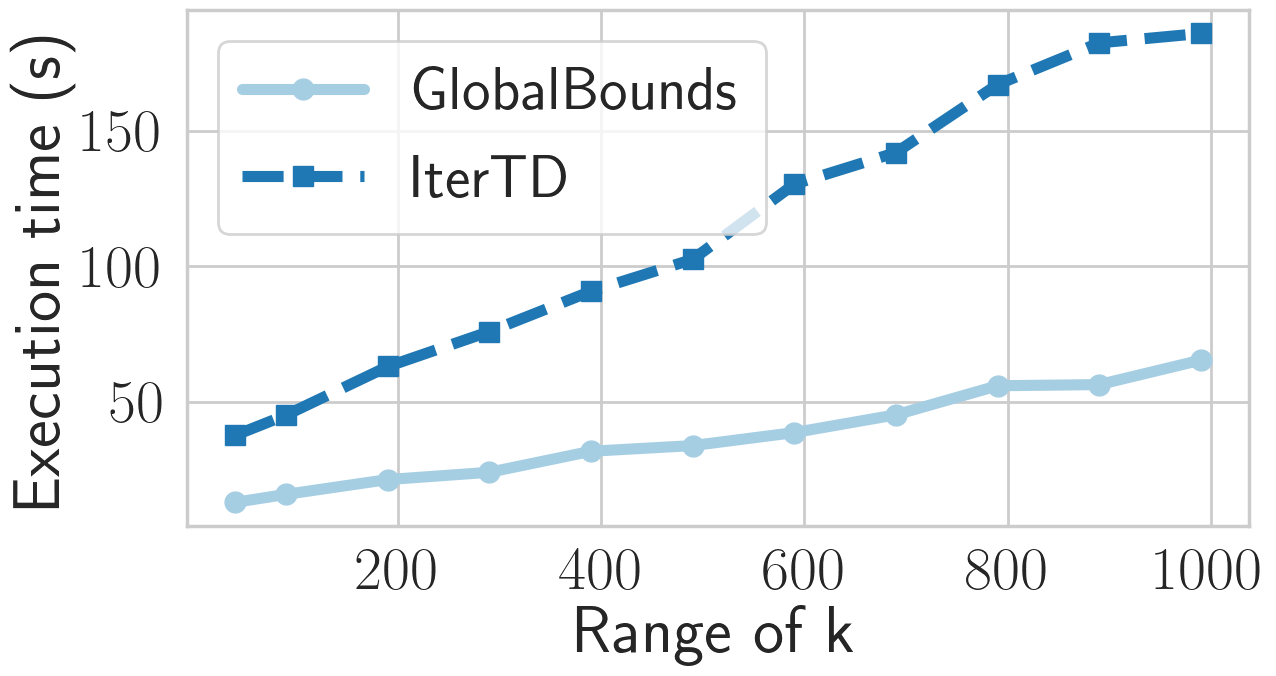}
		\caption{COMPAS dataset}
		\label{fig:compas_k_range_ranking}
	\end{subfigure}%
	\begin{subfigure}[t]{0.3\textwidth}
		\includegraphics[width =0.9\linewidth]{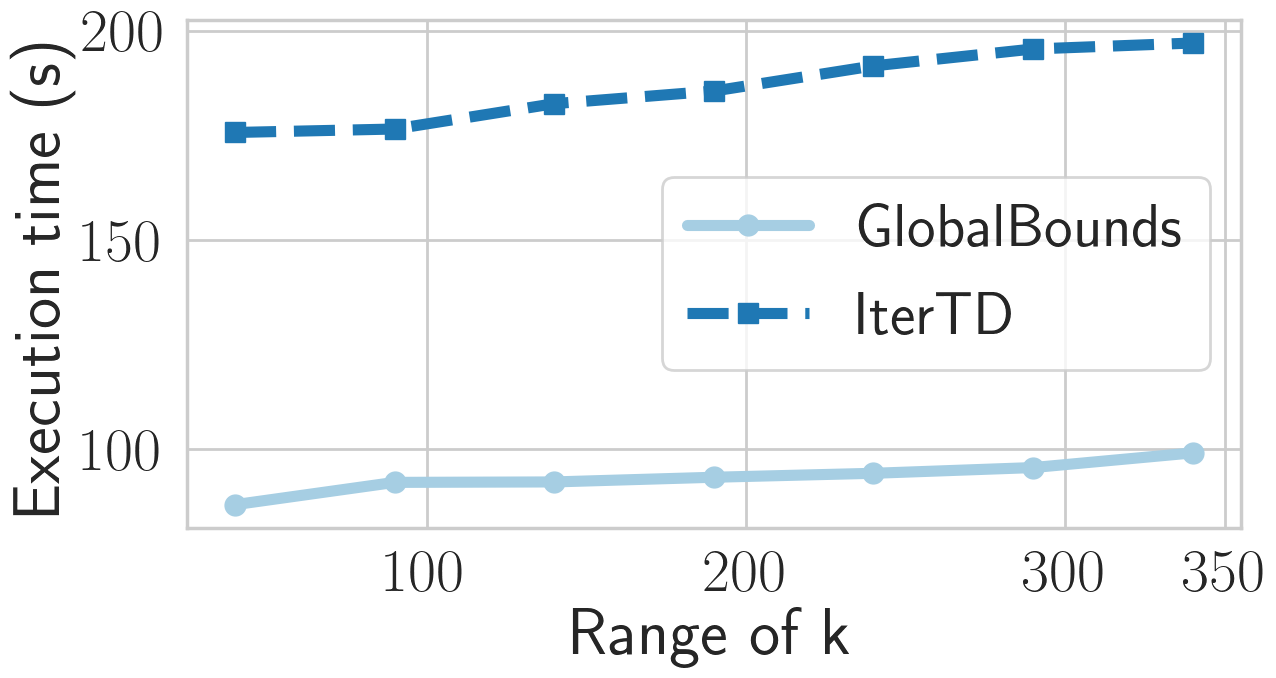}
		\caption{Students dataset}
		\label{fig:students_k_range_ranking}
	\end{subfigure}%
	\begin{subfigure}[t]{0.3\textwidth}
		\includegraphics[width =0.9\linewidth]{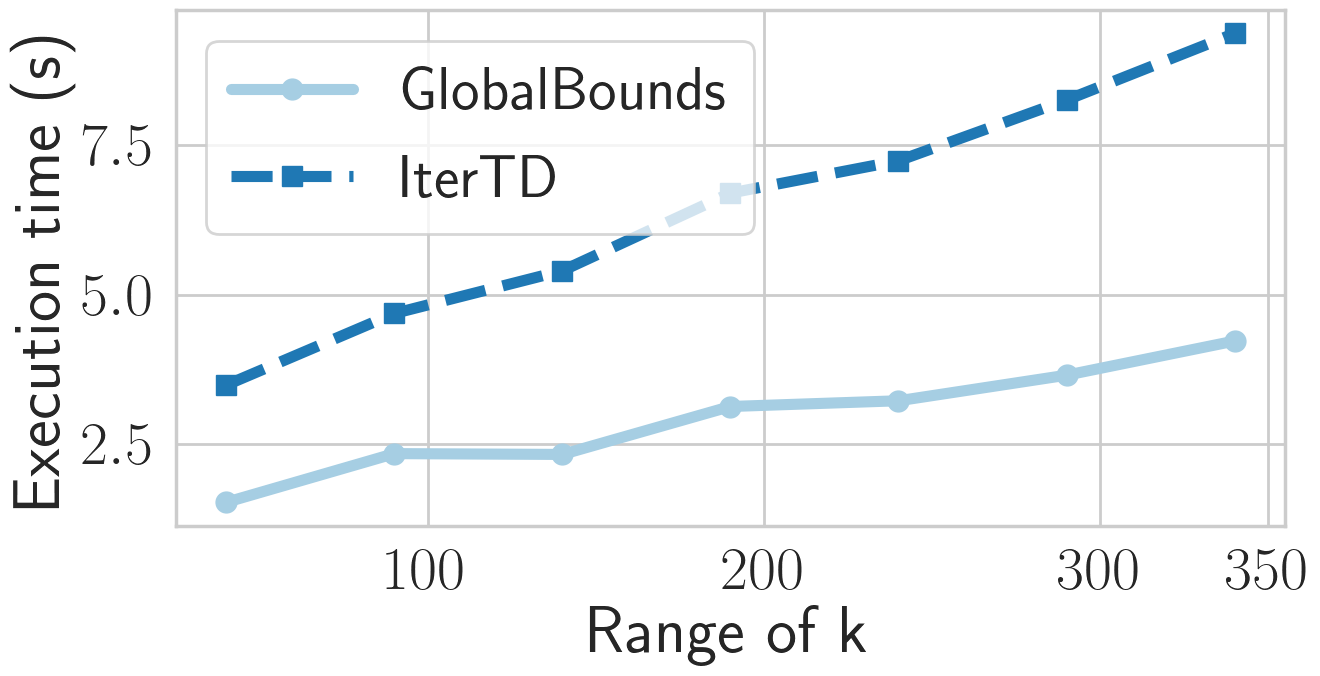}
		\caption{German Credit}
		\label{fig:german_k_range_ranking}
	\end{subfigure}%
	\caption{Running time as a function of the range of $k$- Ranking with global bounds} \label{fig:k_range_ranking}
\end{figure*}


\begin{figure*}
	\centering
	\begin{subfigure}[t]{0.3\textwidth}
		\includegraphics[width = 0.9\linewidth]{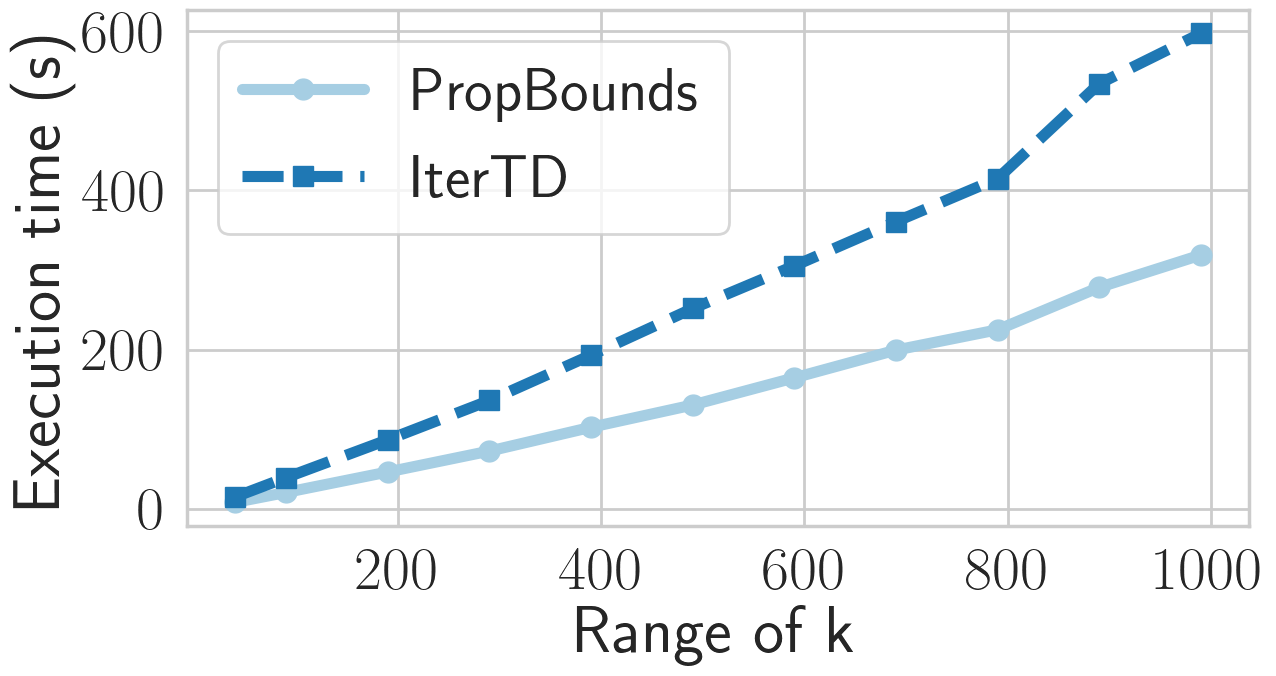}
		\caption{COMPAS dataset}
		\label{fig:compas_k_range_ranking_proportional}
	\end{subfigure}%
	\begin{subfigure}[t]{0.3\textwidth}
		\includegraphics[width =0.9\linewidth]{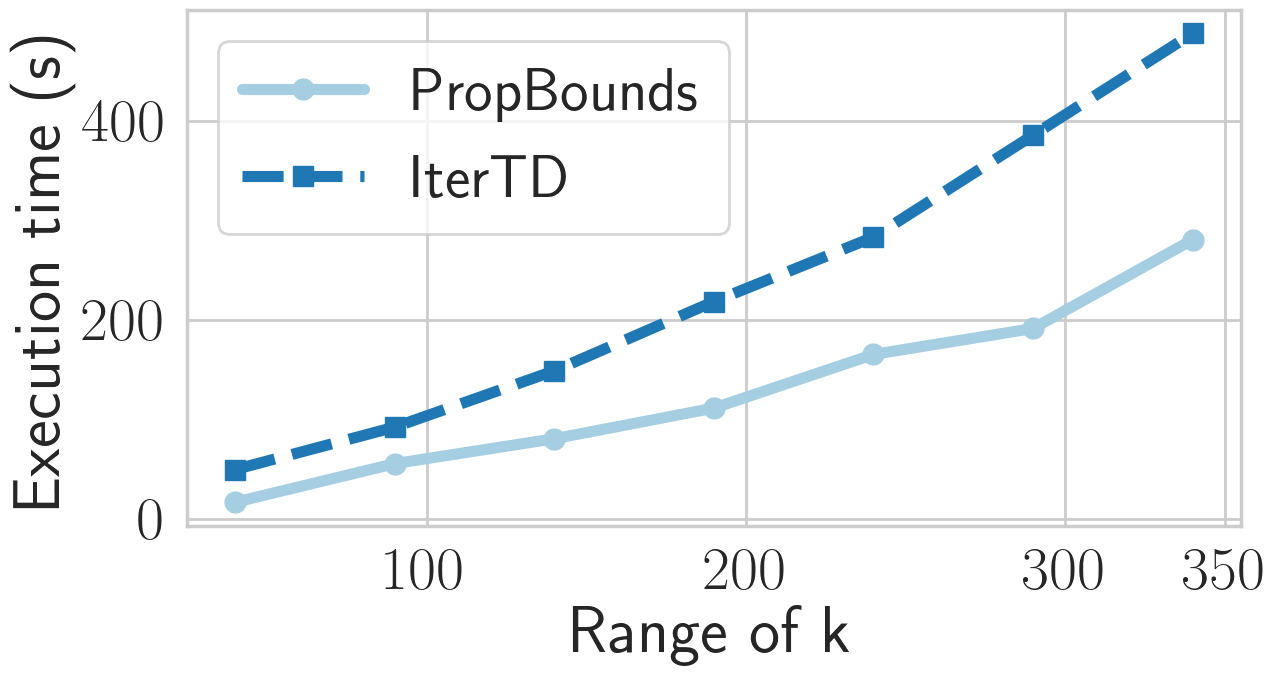}
		\caption{Students dataset}
		\label{fig:students_k_range_ranking_proportional}
	\end{subfigure}%
	\begin{subfigure}[t]{0.3\textwidth}
		\includegraphics[width =0.9\linewidth]{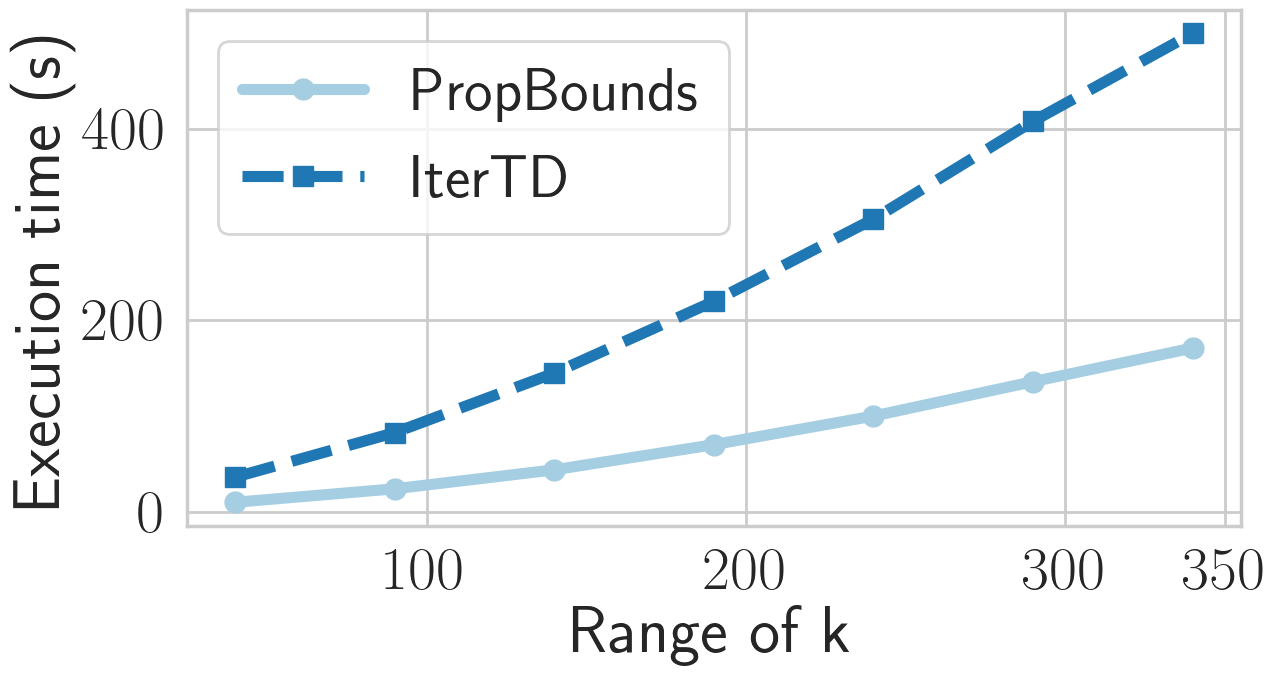}
		\caption{German Credit}
		\label{fig:german_k_range_ranking_proportional}
	\end{subfigure}%
	\caption{Running time as a function of the range of $k$ - Ranking with proportional representation} \label{fig:k_range_ranking_proportional}
\end{figure*}


\paragraph*{Range of $k$} 
We examine the scalability with respect to the range of $k$ considered by the algorithm. We varied the range from 40 (40) to 990 (340) by setting $k_{min}$ to be 10, and increasing $k_{max}$ from 50 (50) to 1000 (350) for COMPAS (Student and German Credit) dataset and observed the effect on the running time.
We set different maximum ranges of $k$ due to the different sizes of the datasets (6889 for COMPAS, 395 for Student, and 1000 for the German Credit). The results are presented in Figure~\ref{fig:k_range_ranking} and \ref{fig:k_range_ranking_proportional}. 
In all cases, the optimized algorithms outperform \textsc{IterTD}, which illustrates the efficient reduction in the search space.

Recall that \textsc{GlobalBounds} and \textsc{PropBounds} optimize the search space compared to \textsc{IterTD} by utilizing the search result of the iteration for $k$ in order to compute the result set for $k+1$. Thus, as the range of $k$ increases, we expect to see a greater improvement in the performance of the optimized algorithm compared to the baseline solution. This trend is shown in Figure~\ref{fig:k_range_ranking} and \ref{fig:k_range_ranking_proportional}. 
To further demonstrate the usefulness of the approach, we compared the number of patterns examined during the search for each one of the algorithms. 
The observed gain was up to 39.35\% in the COMPAS dataset, 56.87\% in the student dataset and 29.27\% in the credit card dataset for detecting groups with biased representation using global bounds, and 39.60\%, 20.49\% and 56.83\% respectively for proportional representation. 



\begin{figure*}
\centering
\begin{subfigure}[t]{0.3\textwidth}
    \centering
    \includegraphics[width=0.9\linewidth]{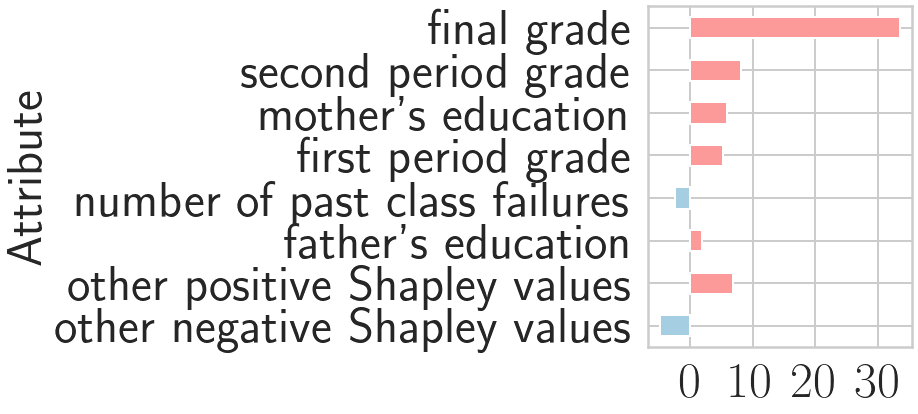}
    \caption{Aggregated Shapley value of group $p_1 = $\{mother's education = primary education\} in the Student datase}
    \label{exp:shap_student_shap}
\end{subfigure}
\hfill
\begin{subfigure}[t]{0.3\textwidth}
    \centering
    \includegraphics[width=0.9\linewidth]{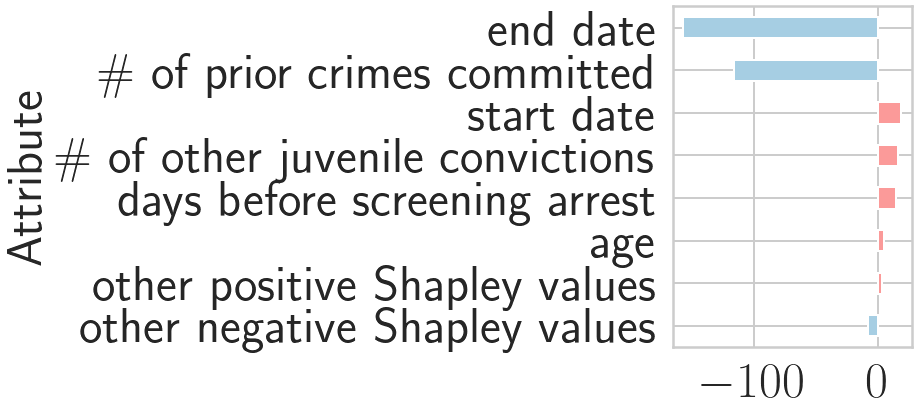}  
    \caption{Aggregated Shapley value of group $p_2 = $\{age = younger than 35\} in the COMPAS dataset }
    \label{exp:shap_compas_shap}
\end{subfigure}
\hfill
\begin{subfigure}[t]{0.3\textwidth}
    \centering
    \includegraphics[width=0.9\linewidth]{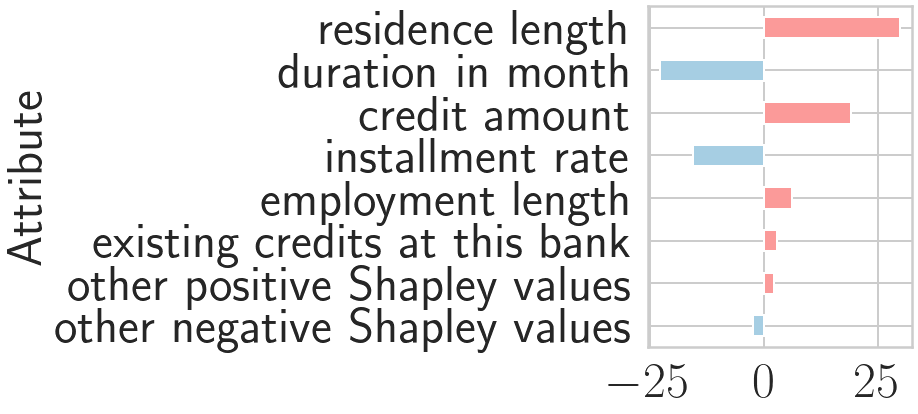}
    \caption{Aggregated  Shapley value of group $p_3 = $\{status of existing account = ($0 \leqslant \dots < 200$DM)\} in the German Credit dataset}
    \label{exp:shap_german_shap}
\end{subfigure}
\hfill
\begin{subfigure}[t]{0.3\textwidth}
    \centering
    \includegraphics[width=0.9\linewidth]{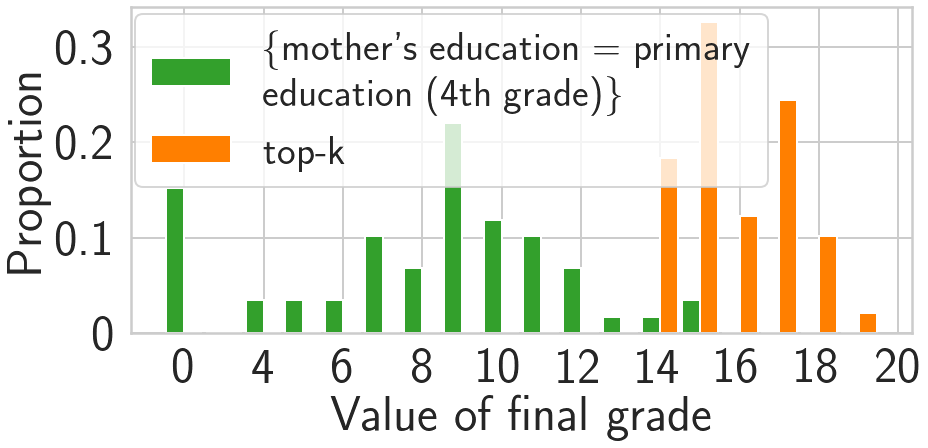}
    \caption{Value distribution of the final grade attribute in the Student dataset}
    \label{exp:shap_student_dis}
\end{subfigure}
\hfill
\begin{subfigure}[t]{0.3\textwidth}
    \centering
    \includegraphics[width=0.9\linewidth]{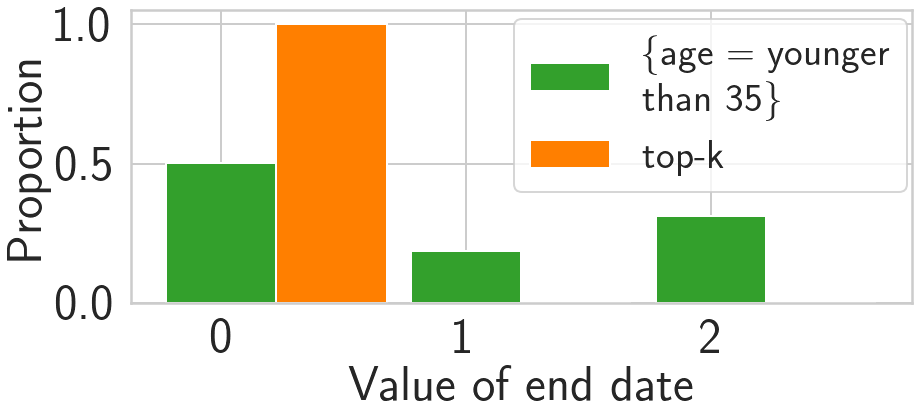}
    \caption{Value distribution of the end date attribute in the COMPAS dataset}
    \label{exp:shap_compas_dis}
\end{subfigure}
\hfill
\begin{subfigure}[t]{0.3\textwidth}
    \centering
    \includegraphics[width=0.9\linewidth]{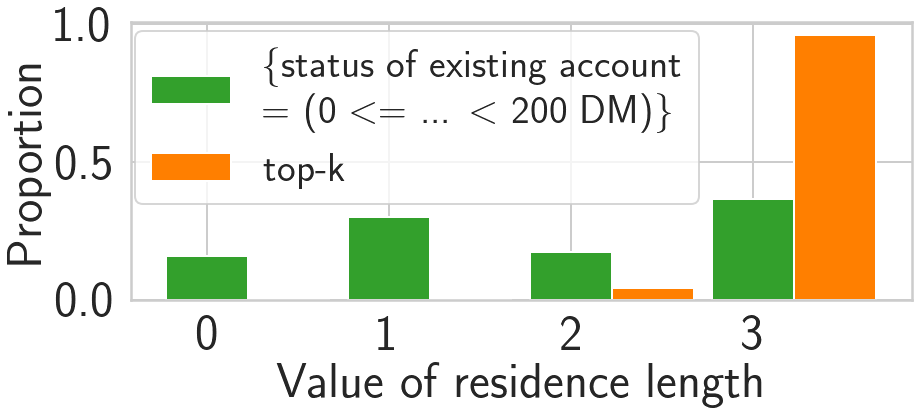}
    \caption{Value distribution of residence length in the German Credit dataset, }
    \label{exp:shap_german_dis}
\end{subfigure}
\caption{Result analysis using Shapley values}
\vspace{-3mm}
\label{exp:shapley_value}
\end{figure*}

\subsection{Result Analysis using Shapley values}\label{sec:shapley_exp}
We next demonstrate the usefulness of our proposed method for results analysis using Shapley values presented in Section~\ref{sec:explanations}. The goal of the experiment is twofold. First, we show that our Shapley values based method for evaluating the effects of attributes on the ranking can indeed reveal useful information on the actual attributes used for ranking when the ranking algorithm is given as a black box. Then we show that the value distribution for those attributes can be used to explain the representation bias, by comparing the distributions for the values in the top-$k$ with those in the detected groups and focusing on the differences.

We trained a regression model using the ranked data for each dataset and examined the Shapley values for groups detected by our algorithms. We present the results for the patterns (groups)  $p_1=\{$mother's education $=$ primary education (4th grade)$\}$ in the Student dataset, $p_2=\{$age $=$ younger than 35$\}$ in COMPAS and $p_3=\{$status of existing account $= (0 \leqslant \dots < 200)$ DM\footnote{A Debit Memo (DM) on a company's bank statement refers to a deduction by the bank from the company's bank account. In other words, a bank debit memo reduces the bank account balance similar to a check drawn on the bank account.}$\}$ from the German Credit dataset, which was detected by the \textsc{GlobalBounds} algorithm for $k=49$ and $L_k = 40$. We observed similar results for other groups detected by the algorithms and other parameters. Figures \ref{exp:shap_student_shap}, \ref{exp:shap_compas_shap} and \ref{exp:shap_german_shap} show the resulting aggregated Shapley values for each group, as explained in Section~\ref{sec:explanations}. We show the Shapley values for the six attributes with the larges values for each group, as the rest had significantly lower values (lower than 5.79\%, 3.31\%, and 9.54\% of the largest aggregated Shapley values for $p_1$, $p_2$ and $p_3$ respectively).




For the group of students whose mother's education level is primary education, which was detected by our algorithm as a group with biased representation in the Student dataset, the final grade has the largest aggregated Shapley value on the ranking (Figure \ref{exp:shap_student_shap}). This result agrees with the fact that the value of the final grade is indeed used for ranking by the ranking algorithm (and it is in fact the only attribute used). Other than the final grade, the first and second period grades have a notable aggregated Shapley (although significantly lower aggregated Shapley value than the final grade). This is due to the high correlation between those attributes and the final grade~\cite{cortez2008using}. We also noticed the mother's education attribute in the result. This may indicate some correlation between the mother's education and the final grade. However, we also note that the aggregation of the Shapley values for other attributes, e.g., father's education, show no clear pattern: some values have a positive effect and some negative, and different tuples in the group have different values. In contrast, all the tuples in the group have the same value (primary education) for the mother's education attribute (since it is used to define the group). Therefore the Shapley values of the attribute for the different tuples in the groups are similar. This may also increase the aggregated value compared to other attributes. We observed this phenomenon, where the attributes used to define the detected group are slightly higher, for other groups detected by our algorithms also. 

For the COMPAS dataset, tuples are ranked by a combined score based on seven attributes: days from compas, the number of other juvenile convictions, days before screening arrest, start date, end date, age, and the number of priors crimes committed.
In Figure \ref{exp:shap_compas_shap}, showing the aggregated Shapley values of people younger than $35$, six out of the above seven attributes are the six attributes with the largest Shapley values. In this case, the attribute end date and the number of priors crimes committed are identified as the most significant factor affecting the detected group.

For the German Credit dataset, tuples are ranked according to their ranking in~\cite{YangS17}, however, the actual ranking algorithm is unknown. Namely, we do not have the ground truth and cannot verify the attribute detected as significant for explaining the bias in the representation of the detected group in the top-$k$ are actually used by the ranking algorithm. The attributes residence length, duration in month, credit amount, and installment rate have the largest Shapley values as shown in Figure~\ref{exp:shap_german_shap}. All of these attributes represent reasonable features to decide one's creditworthiness.

The Shapley value represents the effect of different attributes on the ranking of groups. 
To analyze the differences between the detected groups and top-$k$ tuples (with respect to these attributes), we visualize the value distribution of attributes with the largest Shapley values in Figures~\ref{exp:shap_student_dis},~\ref{exp:shap_compas_dis}, and~\ref{exp:shap_german_dis}. Since the number of tuples in the top-$k$ and the detected group differ, the $y$-axis represents the proportion of tuples (rather than their count) with the values shown on the $x$-axis (the set of possible values for the attribute).

For all three datasets, we observed vast differences in the distributions of the values of the attribute with the largest Shapley value between the tuples in the top-$k$ and the tuples in the group detected with biased representation. For the student dataset (Figure~\ref{exp:shap_student_dis}), the final grades of tuples in the top-$k$ all fall in the range of $15-20$, while most tuples in the detected group have a final grade lower than $15$. In the COMPAS dataset (Figure~\ref{exp:shap_compas_dis}), the value of the end date for all top-$k$ tuples is $0$ while only half of the tuples in the detected group have the same value, and almost $30\%$ of them have the value of $2$. Similar results were observed in the value distribution of the residence length attribute as shown in Figure~\ref{exp:shap_german_dis}.

\subsection{Comparison with Existing Solution}\label{sec:comp}
The problem of identifying subgroups in the data that behave differently compared to the overall dataset was studied in~\cite{pastor2021identifying}. Different from our problem definition, which relies on fairness measures for ranking to define groups with biased representation in the top-$k$ ranked items, the work of~\cite{pastor2021identifying} uses the notion of divergence to measure performance differences among data subgroups. Each data item in the data $t\in D$ is associated with an outcome $o(t)$ where the outcome function is defined based on the ranking of $t$ by the ranking algorithm. The outcome of a group $o(G)$, is then the average of the outcome of every item $t\in G$. The divergence of a subgroup $G$ in the data $D$ is the difference between the outcome of $G$ and the entire data, i.e., $o(G)-o(D)$. Given a threshold $s$ on the subgroup size, the solution of~\cite{pastor2021identifying} computes the divergence of \emph{all} subgroups with sizes larger than $s$.

To better demonstrate the differences between the definitions and the resulting groups identified by each algorithm, we conducted an experiment using the Student dataset. We used the default size threshold of $\tau_s=50$ (support in~\cite{pastor2021identifying} of 0.13, i.e., 13\% of the data). Since~\cite{pastor2021identifying} does not consider a range of $k$'s, we fixed $k_{min}= k_{max}=10$ (namely, compare the results when $k=10$). To allow for easy comparison, we used only the first $4$ attributes of the data: school, sex, age, and address. We used the outcome function $o(t)$ that assigns the value $1$ for tuples $t$ in the top-$k$, and $0$ for the rest (as presented in~\cite{pastor2021identifying}).
Finally, for our algorithms, we used the default parameters of lower bound $10$ for ranking with global bounds and $\alpha = 0.8$ for proportional representation. 

\textsc{PropBounds} outputs $2$ patterns: \{sex=F\} and \{address=R\}, both returned by \textsc{GlobalBounds} as well. Additionally, \textsc{GlobalBounds} returned the patterns \{school=GP\}, \{sex=M\} and \{address=U\}, which had less than 10 instances in the top-$10$ ranked items ($9$, $7$, and $9$ respectively), but considering their overall size ($349$, $208$ and $307$ respectively), their representation in the top-$k$ is adequate and thus are not returned by \textsc{PropBounds}. \rev{The algorithm of~\cite{pastor2021identifying} returned $28$ groups including the groups detected by our solution. Since the number of reported groups may be extremely large, the algorithm of~\cite{pastor2021identifying} ranks the groups by their divergence.} The $5$ patterns with the highest divergence contain $3~-~5$ attributes, with the value assignment sex=M, i.e., they are descendents of the pattern \{sex=M\} (in the pattern graph) returned by \textsc{GlobalBounds}. The pattern \{sex=M\} was ranked at $17$ according to its divergence value.



\rev{
Our algorithms differ from the solution of~\cite{pastor2021identifying} mainly in how they define the target groups.
The two solutions deal with a similar problem, however, ours prefers concise groups (most general pattern) while the solution of~\cite{pastor2021identifying} is designed to identify \emph{all} groups with sufficient representation in the overall data and high divergence (a measure of ``unfairness''). Consequently, the output of~\cite{pastor2021identifying} is typically larger and contains subgroups that are consumed by each other. Finally, the work of~\cite{pastor2021identifying} considers a single $k$ while we consider a range of $k$'s, aligning with fairness definitions in the literature, making the solution fair for any position in the top-$k$.}


%% file: related.tex
\section{Related Work}\label{sec:related}
The notion of fairness in ranking algorithms was studied in a line of works, introducing different fairness definitions~\cite{CelisSV18, YangS17, jarvelin2002cumulated, zehlike2017fa, SinghJ18, rankingFairness}. These definitions typically focus on top-$k$ positions, as those are usually the most important positions. In this paper, we consider two such definitions: the fundamental definition of~\cite{CelisSV18}, which measures fairness by bounding the representation of different groups in the data, and a refined definition that considers proportional groups representation. These definitions, as customary in the context of algorithmic fairness, refer to some given protected groups. We harness these definitions to define the problem of detecting groups with biased representation, eliminating the need to pre-define protected groups. The problem of generating fair ranking results was studied in~\cite{zehlike2017fa, asudeh2019designing}.
These works consider a wide range of definitions for fairness in ranking, which rely on the notion of  protected groups. This line of work is orthogonal to the problem we defined in this paper, and our proposed method can be used to identify such protected groups, when they are unknown in advance.





Recent works have studied the problem of automatically detecting ``problematic'' or biased subgroups in the data, without the need to specify the protected attributes a priori, in the context of classification~\cite{ChungKPTW20, cabrera2019fairvis, pastor2021looking, pastor2021identifying}. 
In \cite{pastor2021looking}, the authors introduced the notion of divergence to measure the difference in the behavior of a classifier on data subgroups. The goal is then to report subgroups with sizes above a given threshold and high divergence. In~\cite{pastor2021identifying} they extend their framework to ranking, where they consider the average outcome value, which is defined based on the ranking of the instances in each group, as a measure of the group's outcome. In contrast to~\cite{pastor2021identifying}, our problem definitions rely on groups' representation in the top-$k$ ranked items as fairness measures for ranking. We demonstrate the differences in the resulting groups identified by each definition in Secetion~\ref{sec:comp}.
The interactive system MithraCoverage~\cite{jin2020mithracoverage} investigates population bias in intersectional groups. The notion of coverage is introduced to identify intersectional subgroups with inadequate representation in the dataset. Differently, in our work, we only report patterns with adequate representation in the data, but inadequate representation in the output of a ranking algorithm.

The use of Shapley values to provide explanations for ML models was studied in a line of works (see e.g.,~\cite{StrumbeljK14, LundbergL17}). In these works, the Shapley values are computed for an \emph{individual} input instance to a \emph{classification or regression model}. The Shapley value of a feature is then interpreted as the contribution of the feature to the output of the given input. Differently, we are aiming at providing explanations for the representation of a \emph{group of tuples} in the output of a \emph{ranking algorithm}. In~\cite{pastor2021looking} the authors presented a method to measure the contribution of items to the divergence of groups utilizing Shapley values. However, it considers only the contribution of attributes that are used to define the group. In contrast, our solution considers all attributes as possible explanations. This requires an additional aggregation step in the computation of the Shapley values. As demonstrated in~\ref{sec:shapley_exp}, the explanation is typically buried in the values of attributes used for ranking. Moreover, our adjustment of Shapley values to explain a ranking algorithm is novel.

Our baseline solution, utilizing a top-town search presented in Section~\ref{sec:top-down} is built on the algorithm presented in~\cite{AsudehJJ19} (for a simpler problem), which in turn shares similar ideas to the Apriori algorithm~\cite{AgrawalS94}, the Set-Enumeration Tree for enumerating sets in a best-first fashion~\cite{Rymon92}, discovering functional dependencies (FDs) \cite{apenbrockEMNRZ15, HuhtalaKPT99} and frequent item-sets and association rule mining~\cite{AgrawalS94, zaki2000scalable}.
Similarly to \cite{AsudehJJ19}, the key difference from our work lies in the structure of the graph traversed in the solution: the pattern graph (in our case) compared to the powerset lattice in the other works.
\rev{Conditional functional dependencies (CFDs)~\cite{fan2008conditional} extend the notion of FDs by considering patterns to describe dependencies that hold only on subgroups in the data. Similar to the top-down search applied by the baseline solution, algorithms for discovering CFDs~\cite{fan2010discovering, diallo2012discovering, golab2008generating, rammelaere2019revisiting} also utilize the notion of pattern and lattice of patterns. However, the difference in the end goal (discovering CFDs versus identifying groups with biased representation) leads to differences in the pruning techniques in the baseline solution.} 
We then present two novel optimized algorithms designed for each one of the problems we defined. These algorithms reduce the search space as explained in Section~\ref{sec:algo} and significantly outperform the baseline solution as shown in the experimental evaluation.

%% file: conc.tex
\section{Conclusion}\label{sec:conc}
In this paper, we have studied the problem of detecting groups with biased representation in the result of a ranking algorithm. We build on fairness measures previously defined in the literature, considering the representation of protected groups in the top-$k$ ranked items, for any reasonable range of $k$. Our problem definitions eliminate the need to pre-define the protected groups. We consider two variants of the problem. The first is based on global bounds over the representation of different groups in the top-$k$ ranked items, and the second restricts the representation of each group in the top-$k$, based on its overall representation in the data. 

We theoretically analyse the complexity of the problem, showing that in the worst case, the number of groups can be exponential in the number of the dataset attributes. We present a baseline algorithm that can handle both definitions and two optimized algorithms designed to improve the performance for each fairness measure. Furthermore, we present a method to explain the output of our algorithms. There are many intriguing directions for future research, including the extension of the framework to support other fairness measures and further investigation of the automatic suggestion for thresholds.

